\documentclass[journal]{IEEEtran}

\ifCLASSINFOpdf

\else

\fi

\usepackage{framed}
\usepackage[ruled]{algorithm2e}
\usepackage{amssymb}
\usepackage{latexsym}
\usepackage{epsfig}
\usepackage{float}
\usepackage{booktabs,multirow, multicol}
\usepackage{cite}
\usepackage{rotating}
\usepackage{epstopdf}
\usepackage[margin=1in]{geometry}
\usepackage{graphicx}
\usepackage[pagebackref=false,breaklinks=true,colorlinks,bookmarks=false]{hyperref}
\usepackage{color}
\usepackage{amsmath}
\usepackage{verbatim}
\usepackage{gensymb}
\usepackage{array, longtable}
\usepackage{wrapfig}
\usepackage{enumitem}
\usepackage{tabularx}
\usepackage{url}
\usepackage[table]{xcolor}
\usepackage{tikz}
\usepackage[altpo]{backnaur} 
\usepackage{epsfig}
\usepackage{graphicx}
\usepackage{amsmath}
\usepackage{amssymb}
\usepackage{gensymb}
\usepackage{footmisc}

\usepackage{cite}
\usepackage{array}
\usepackage{url}

\hypersetup{colorlinks,breaklinks,
            urlcolor=[rgb]{0,0.0,0.0},
            linkcolor=[rgb]{0,0.0,0.0},
            citecolor=[rgb]{0,0.0,0.0}}

\newcommand*{\MinNumber}{-0.01}%
\newcommand*{\MidNumber}{50.1} %
\newcommand*{\MaxNumber}{100.0}%
\newcommand*{\LowNumber}{5.0}%

\usepackage{pifont}% http://ctan.org/pkg/pifont

\newcommand{\sdt}{PDT{} }

\newcommand{\proposed}{Prepended Domain Transformer{} }

\usepackage{collcell}
\usepackage{xcolor}
\usepackage{colortbl}
\newcommand{\AGO}[1]{%

        \ifdim #1 pt > \MidNumber pt
            \pgfmathsetmacro{\PercentColor}{max(min(100.0*(#1 - \MidNumber)/(\MaxNumber-\MidNumber),100.0),0.00)} %
            \textcolor{green!\PercentColor!blue}{#1}

        \fi

        \ifdim #1 pt < \MidNumber pt
        
            \ifdim #1 pt > \LowNumber pt
            
                  \pgfmathsetmacro{\PercentColor}{max(min(100.0*(\MidNumber - #1)/(\MidNumber-\MinNumber),100.0),0.00)} %
                \textcolor{orange!\PercentColor!green}{\textbf{#1}}
                
                \else
                
            \pgfmathsetmacro{\PercentColor}{max(min(100.0*(#1 - \LowNumber)/(\MinNumber-\LowNumber),100.0),0.00)} %
            \textcolor{red!\PercentColor!orange}{\textbf{#1}}
            
        \fi
        
        \fi

}

\newcommand{\AG}[1]{%

        \ifdim #1 pt > \MinNumber pt
            \textcolor{black}{#1}

        \fi

}

\graphicspath{{pics/}}

\usepackage{amsmath}
\usepackage{makecell}
\usepackage{multirow}
\usepackage{array}
\usepackage{longtable}
\usepackage{tabularx}
\usepackage{arydshln}

\usepackage{siunitx}
\usepackage{booktabs}
\definecolor{Gray}{gray}{0.94} %0.94

\newcommand{\etal}{\textit{et al.} }

\newcolumntype{L}{>{$}l<{$}}
\newcolumntype{C}{>{$}c<{$}}
\newcolumntype{R}{>{$}r<{$}}

\newcommand{\nm}[1]{\textnormal{#1}}

\begin{document}

\title{Prepended Domain Transformer: Heterogeneous Face Recognition without Bells and Whistles}

\author{Anjith~George, Amir Mohammadi and~Sebastien~Marcel% <-this % stops a space
\thanks{A. George, A. Mohammadi and S. Marcel are in Idiap Research Institute, Centre du Parc, Rue Marconi 19, CH - 1920, Martigny, Switzerland \\
(e-mail: {anjith.george,amir.mohammadi,sebastien.marcel}@idiap.ch).}% <-this % stops a space
%\thanks{This work was supported by the BATL project funded by IARPA.}
\thanks{Manuscript received September xx, 2021; revised September xx, 2021.}}

\markboth{IEEE TRANSACTIONS ON INFORMATION FORENSICS AND SECURITY}%
{}

\maketitle

\begin{abstract}

Heterogeneous Face Recognition (\textit{HFR}) refers to matching face images captured in different domains, such as thermal to visible images (VIS), sketches to visible images, near-infrared to visible, and so on. This is particularly useful in matching visible spectrum images to images captured from other modalities. Though highly useful, \textit{HFR} is challenging because of the domain gap between the source and target domain. Often, large-scale paired heterogeneous face image datasets are absent, preventing training models specifically for the heterogeneous task. In this work, we propose a surprisingly simple, yet, very effective method for matching face images across different sensing modalities. The core idea of the proposed approach is to add a novel neural network block called \proposed (PDT) in front of a pre-trained face recognition (FR) model to address the domain gap. Retraining this new block with few paired samples in a contrastive learning setup was enough to achieve state-of-the-art performance in many \textit{HFR} benchmarks. The \sdt blocks can be retrained for several source-target combinations using the proposed general framework. The proposed approach is architecture agnostic, meaning they can be added to any pre-trained FR models. Further, the approach is modular and the new block can be trained with a minimal set of paired samples, making it much easier for practical deployment. The source code and protocols will be made available publicly.

\end{abstract}


\IEEEpeerreviewmaketitle

\section{Introduction}
%============================================================================================================================================================

\begin{figure}[t!]
  \centering
  \includegraphics[width=0.7\linewidth]{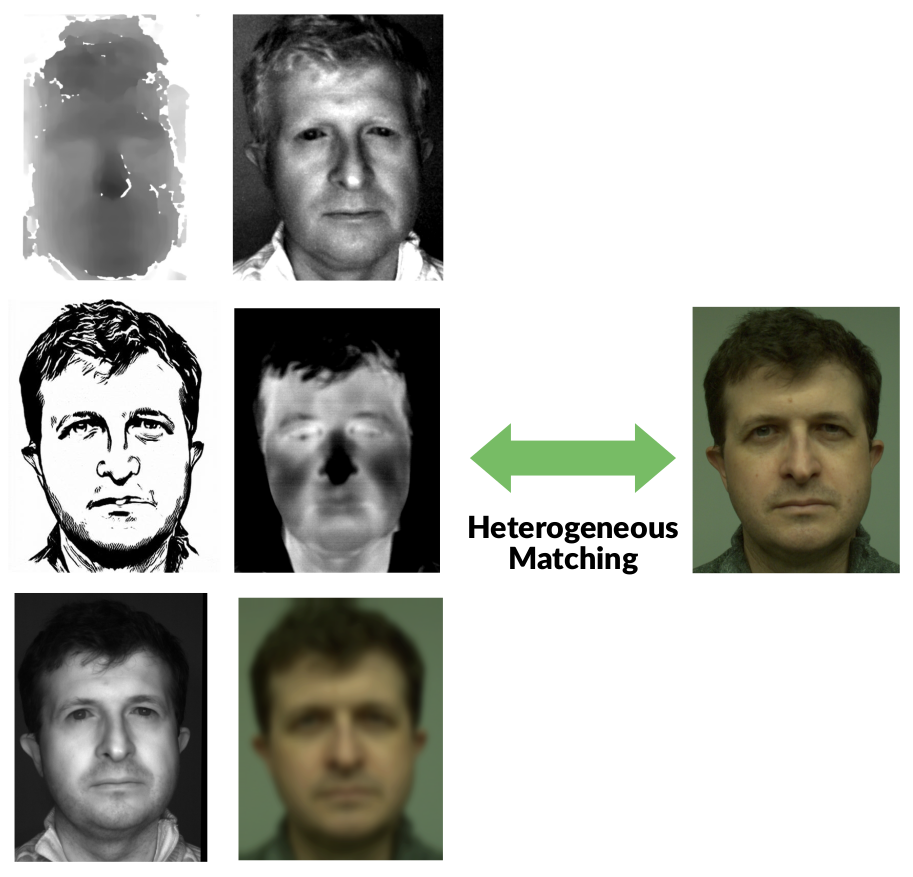}
  \caption{This figure shows the face of the same person captured with several different modalities such as depth, short-wave infrared, sketch, thermal, near-infrared and blurred faces. The task in \textit{HFR} is to perform cross-modal face recognition given the RGB reference images and the probe image from the new modalities.}  \label{fig:hfr}
\end{figure}

\IEEEPARstart{F}{ace} recognition (FR) systems offer a convenient way for access control. Most of the state-of-the-art FR methods achieve excellent performance in `in the wild' conditions and human parity in face recognition performance \cite{learned2016labeled}, thanks to convolutional neural networks. Typical FR systems operate in the homogeneous domain, meaning the enrollment and matching are performed with the same modality, typically with face images obtained from an RGB camera (visible spectrum). However, in several scenarios, matching in a heterogeneous setting could be advantageous. For example, near-infrared (NIR) cameras, ubiquitous in mobile phones and surveillance cameras, offer superior performance irrespective of the illumination conditions \cite{li2007illumination}. However, an FR system operating in a homogeneous setting will require enrollment samples obtained under the same NIR camera. Heterogeneous face recognition (\textit{HFR}) systems try to alleviate this limitation, by enabling ways to perform cross-domain matching. For example, with an \textit{HFR} system, enrolled RGB images could be matched with NIR images, obviating the requirement for enrollment of different modalities \cite{klare2012heterogeneous}. This also opens up the possibility of using additional channels for recognition, such as thermal and shortwave-infrared, without having enrollment samples in the corresponding modalities.
Thermal images can be acquired without using active illumination and hence could be used robustly in day and night conditions. In short, the \textit{HFR} task is very useful in practical applications and could provide a way to extend FR to challenging and uncontrolled environments.

Though \textit{HFR} is very useful, it is very challenging for several reasons. First, there is a large domain gap between the images captured by various sensors. The performance of the networks trained using RGB images deteriorates when used with images captured with other sensing modalities \cite{he2018wasserstein}. Further, there are not many large-scale heterogeneous datasets to train models specifically for \textit{HFR} \cite{poster2021large}. Leveraging pre-trained FR models, which are trained using large-scale face recognition datasets, appears to be a reasonable strategy to follow in this limited data domain. Furthermore, special emphasis should be put on developing approaches that work with a limited amount of paired data as this type of heterogeneous data is costly to acquire and not readily available \cite{poster2021large}. 

In this work, we propose a novel, yet surprisingly simple approach for \textit{HFR}. We leverage a pre-trained FR model as one of the key components in our framework. Our approach doesn't depend on the selection of architecture of the FR model giving it maximum flexibility in deployment. We achieve this by prepending a new network module, called \proposed (PDT), to a pre-trained FR module to transform the target domain images. The only learnable component is the new prepended module, which is very parameter efficient and obtains excellent performance with few paired samples. This method is very practical in deployment scenarios since one just needs to prepend a new module to convert a typical FR pipeline to an \textit{HFR} pipeline. The approach is generic and can be retrained easily for any pair of heterogeneous modalities. Through extensive evaluations, we show that this simple addition achieves state-of-the-art results in many challenging \textit{HFR} datasets. The framework's design is intentionally kept simple to demonstrate the effectiveness of the approach and to allow for future extensions. Moreover, the parameter and computational overhead added by the framework is negligible, making the proposed approach suitable for real-time deployment.

The main contributions of this work are listed below:

\begin{itemize}
\item We propose a heterogeneous face recognition framework, leveraging a pre-trained FR model together with a \proposed block. The approach is architecture agnostic and can be used in many heterogeneous recognition scenarios. The method has minimal computational overhead and a very minimal number of learnable parameters, which makes it easy for deployment.

\item We perform an extensive set of experiments on public datasets and validate the effectiveness of the proposed approach in different heterogeneous face recognition settings.

\item We introduce a new multi-channel heterogeneous face recognition (MCXFace) dataset which consists of both homogeneous and heterogeneous protocols with channels such as color, thermal, depth, stereo, infrared, short-wave infrared, and synthesized 3D maps. We provide standard protocols and baselines for the heterogeneous protocols in this dataset.

\end{itemize}

Finally, the source codes are made available publicly to make it easier to extend the work further \footnote{\url{https://gitlab.idiap.ch/bob/bob.paper.tifs2022_hfr_prepended_domain_transformer}}.

The rest of the paper is organized as follows. Section \ref{sec:related_work} presents recent literature on \textit{HFR}. Details of the proposed approach are described in Section \ref{sec:approach}. Extensive evaluation of the proposed approach, along with comparisons with the state-of-the-art, and discussions, are presented in Section \ref{sec:experiments}. Conclusions and future directions are described in Section \ref{sec:conclusions}.

\section{Related work} 
\label{sec:related_work}
%============================================================================================================================================================ 

The task in \textit{HFR} is matching face images collected in different sensing modalities. The challenge lies in the fact that the image of the same subject appears very different in different modalities (domain gap). This difference in appearance increases the within-class variance making matching difficult, and a direct comparison of these heterogeneous images degrade the performance. Several approaches have been proposed in the literature to address this limitation.

\subsection{Common-space projection method}
%--------------------------------------------

Common-space projection methods \cite{kan2015multi,he2017learning} aim to learn a mapping to project different face modalities into a common shared subspace in an effort to reduce the domain gap. Lin and Tang \cite{lin2006inter} developed a common discriminant feature extraction method for extracting features from cross-modal images and projecting them onto a common feature space. Canonical correlation analysis (CCA) was proposed as a way to match face images between NIR and VIS by Yi \etal \cite{yi2007face}. To learn mapping functions that connect cross-modality domains and common spaces, regression-based approaches were proposed by authors in \cite{lei2009coupled,lei2012coupled}. Sharma and Jacobs \cite{sharma2011bypassing} proposed a partial least squares-based method to learn a linear mapping for face image across different modalities so that the mutual covariance is maximized. Klare and Jain \cite{klare2012heterogeneous} proposed a way to represent face images in terms of their similarity to a set of prototype face images. The prototype-based face representation was then projected onto a linear discriminant subspace, which was used to perform the recognition.  In \cite{de2018heterogeneous}, the authors proposed a novel approach to the \textit{HFR} task called Domain-Specific Units (DSU). Essentially, they suggest that high-level features of convolutional neural networks trained on the visible spectrum are domain-independent, and they can be used to 
encode images captured in other sensing modalities. Subsequently, they proposed adapting the initial layers (DSUs) of a pre-trained FR model to make it suitable for different heterogeneous scenarios. Essentially, adapting the lower layers reduces the domain gap, and the whole pipeline is trained in a contrastive setting. However, the number of layers that need to be adapted is a hyper-parameter that has to be found with an extensive set of experiments. Furthermore, this approach requires the model to be adapted for each different architecture of pre-trained FR models. Recently, Cheema et al. \cite{cheema2022heterogeneous} proposed a Cross Modality discriminator network (CMDN) for HFR. The architecture of CMDN follows a standard ResNet50 model with the squeeze and excitation module (SENet-50) \cite{hu2018squeeze}. Their approach uses a Deep Relational Discriminator (DRD) module to learn cross-domain matching. This DRD module is essentially a multi-layer perceptron (MLP) supervised by binary cross entropy loss (BCE).
The learning of the CMDN module is supervised by the unit class loss which is a combination of triplet loss and a modified version of triplet loss that uses class means.  The CMDN module (SENet-50) is initialized from a pre-trained face recognition backbone trained on the VGGFace2 dataset \cite{cao2018vggface2}. The CMDN network is further trained on another Visible-Thermal face dataset (IRIS face dataset \cite{zhang2018tv}). The tuned CMDN network is further trained for the HFR task using a variant of triplet loss (Unit class loss). Now the embeddings obtained from two modalities (gallery and probe modalities) are concatenated for positive and negative pairs and an MLP model (DRD module) is trained on top of these concatenated embeddings with the BCE loss. The scoring is performed with probe-gallery pairs, and three different strategies were used for scoring 1) Using the embeddings and cosine loss, 2) using the output of the DRD module, and 3) score fusion of 1 and 2. Their evaluation strategy uses pairs of samples instead of probes against the gallery. They reported that the fusion model achieves better results.

\subsection{Invariant feature based methods}
%--------------------------------------------

Invariant feature-based methods aim to extract modality invariant features to match heterogeneous face images. Liao \etal\cite{liao2009heterogeneous} propose to use Difference of Gaussian (DoG) filters to highlight the structure of the images. Then, they use multi-scale block local binary patterns (MB-LBP)\cite{liaoLearningMultiscaleBlock2007} as features and train a subspace-based face recognition system jointly on the samples of source and target domains. In \cite{klare2010matching}, Klare \etal proposed a local feature-based discriminant analysis (LFDA) for the \textit{HFR} task. They extracted scale-invariant feature transform (SIFT) \cite{lowe2004distinctive} and multi-scale local binary pattern (MLBP) \cite{ojala2002multiresolution}  feature descriptors as a patch unit from the sketch and VIS images for the \textit{HFR} task. Zhang \etal \cite{zhang2011coupled} proposed a coupled information-theoretic encoding (CITE) extraction method to maximize the mutual information between the heterogeneous modalities in the quantized feature spaces. The local maximum quotient (LMQ) was proposed to extract invariant characteristics in cross-modality facial images in \cite{roy2018novel}. Several works have also used convolutional neural network (CNN) based methods \cite{he2017learning,he2018wasserstein} to extract invariant features for the \textit{HFR} task.

\subsection{Synthesis based methods}
%--------------------------------------------

Synthesis-based \textit{HFR} methods \cite{tang2003face,fu2021dvg} attempt to synthesize the source domain (VIS in most of the cases) from the target modality, after synthesizing the source images typical face recognition networks can be used to perform the biometric matching.
Authors in \cite{wang2008face} proposed a patch-based synthesis approach to generate VIS to sketches and reverse using Multi-scale Markov Random Fields. The approach was evaluated using several face recognition methods such as Eigenfaces, Fisherfaces, dual space LDA, and so on. The work in \cite{liu2005nonlinear} used Locally Linear Embedding (LLE) to learn a pixel-level mapping between VIS images and viewed sketches. Authors in \cite{baeNonvisualVisualTranslation2020} use CycleGAN \cite{zhuUnpairedImagetoImageTranslation2017} to transform images from the target domain to the source domain. Contrastive loss using a Siamese network is added during the training of CycleGAN to preserve the identity of faces. Moreover, the images are pre-processed before inputting to CycleGAN using \cite{tanEnhancedLocalTexture2010} to reduce the domain gap further.
Authors in \cite{zhang2017generative}, proposed a Generative Adversarial Network-based Visible Face Synthesis (GAN-VFS) method to synthesize photo-realistic visible face images from polarimetric images. Identity loss was combined with a perceptual loss in the training process. The synthesized visible images were further used by a VGG network to extract the embeddings. Their method was evaluated on the Polathermal dataset and achieved an average Equal Error Rate of 34.58\%. With the advancement in the development of stable methods to train GANs, several recent approaches have been proposed using GANs for the synthesis of VIS images from another modality. The work in \cite{fu2021dvg} treated \textit{HFR} as a dual generation problem and proposed a Dual Variational Generation (DVG-Face) framework. A Dual generator was designed to learn the joint distribution of heterogeneous pairs and to generate heterogeneous pairs to address the lack of adequate data to train the \textit{HFR} model. A pairwise identity preserving loss on the generated images was employed to ensure identity consistency. The generated images are used to train the \textit{HFR} network in a contrastive setting. This approach achieved state-of-the-art results in many challenging \textit{HFR} benchmarks.

\subsection{Limitations of current approaches}
%---------------------------------------------

The majority of recent \textit{HFR} methods proposed in the literature \cite{fu2021dvg,zhang2017generative} utilize synthesis-based methodologies. GAN-based synthesis methods have become popular due to their ability to generate high-quality images combined with breakthroughs in training them \cite{salimans2016improved}. Furthermore, synthesis-based \textit{HFR} approaches can benefit from the use of a pre-trained FR model, which eliminates the requirement to train the model with a huge amount of training data.

However, this technique falls short when it comes to actual use cases. The synthesize-based \textit{HFR} must first generate an RGB image from the target modality image before passing it through a FR model to retrieve the embeddings at inference time. The synthesis process adds a significant amount of computing cost, which may restrict its usefulness in practical deployment scenarios. Moreover, synthesis-based (generative) algorithms are often trained to generate RGB images that are optimized for both perceptual and identity loss \cite{fu2021dvg} (along with several other metrics). However, because the generated pictures are utilized in conjunction with a neural network, the perceptual quality of the generated images may not be significant in the context of the \textit{HFR} task. In other words, it is critical to retain discriminative traits that potentially match those in the source class rather than generating high-fidelity images. This is particularly important as the generation process is often a much harder problem to solve when the amount of paired training data is limited.

\section{Proposed Method}
\label{sec:approach}
%============================================================================================================================================================

We follow the definitions in \cite{weiss2016survey,de2018heterogeneous} to formalize the\textit{HFR} task.

\subsection{Formal definition of \textit{HFR}}
%--------------------------------------

Consider a domain $\mathcal{D}$ with samples $X \in \mathbb{R}^d$ and a marginal distribution $P(X)$ (with dimensionality-$d$). The task of an FR system $\mathcal{T}^{fr}$ can be defined by a label space $Y$ whose conditional probability is $P(Y|X,\Theta)$, where $X$ and $Y$ are random variables and $\Theta$ defines the model parameters. In the training phase of an FR system, $P(Y|X, \Theta)$ is typically learnt in a supervised fashion given a dataset of faces $X=\{x_1, x_2, ..., x_n\}$ together with their identities $Y=\{y_1, y_2, ..., y_n\}$.

Now consider the following heterogeneous face recognition (\textit{HFR}) problem. Here we assume that we have two domains, source domain  $\mathcal{D}^s = \{X^s, P(X^s)\}$ and target domain $\mathcal{D}^t = \{X^t, P(X^t)\}$ sharing the labels $Y$.

In a broad sense, the task in the \textit{HFR} problem $\mathcal{T}^{hfr}$ is to find a $\hat{\Theta}$, where $P(Y|X^s, \Theta) = P(Y|X^t, \hat{\Theta})$. The form and the nature in which $\hat{\Theta}$ is estimated varies with different \textit{HFR} approaches.  % TODO: check this part

\subsection{Proposed approach}

\begin{figure*}[t!]
  \centering
  \includegraphics[width=0.99\linewidth]{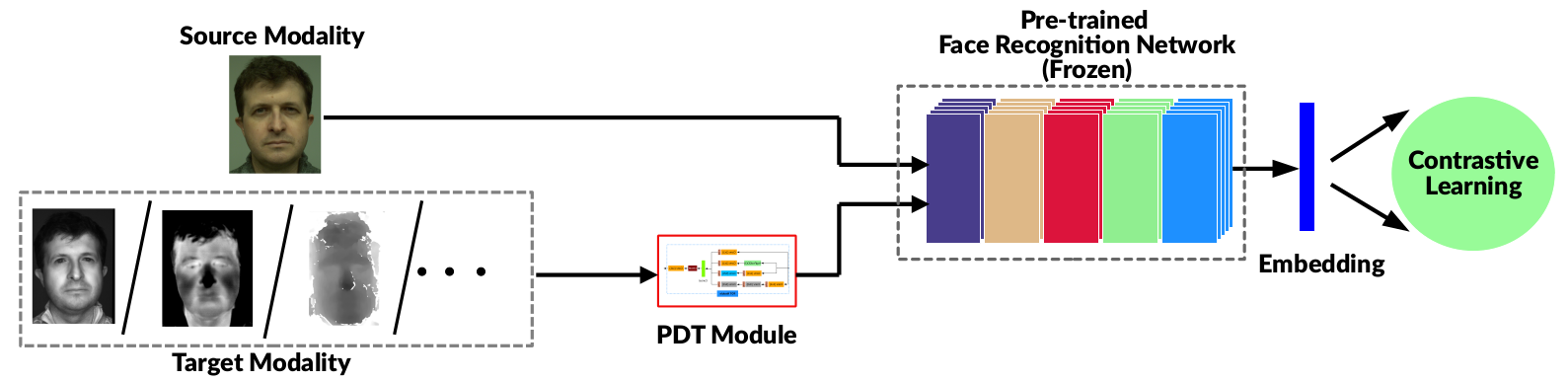}
  \caption{Schematic diagram of the proposed framework. Target domain images are transformed using the proposed \proposed (PDT) block. The \sdt block prepended to the frozen FR model is trained in a standard Siamese setting with contrastive loss.}
  \label{fig:framework}
\end{figure*}

In the proposed approach, let us first assume that the samples from both domains, $X_s=\{x_1, x_2, ..., x_n\}$ and $X_t=\{x_1, x_2, ..., x_n\}$  from $\mathcal{D}^s$ and $\mathcal{D}^t$ with the shared set of labels  $Y=\{y_1, y_2, ..., y_n\}$ are available. Also, assume that the parameters of an FR model $\Theta$ (we denote it as $\Theta_{FR}$ in the following discussion) for the  (VIS) model is available from $\mathcal{D}^s$. In our case, $\Theta_{FR}$ is essentially the parameters of a pre-trained FR model trained using visible spectrum images. Following the discussion on the synthesis based HFR, we hypothesize that a module with a learnable set of parameters $\theta_{PDT}$ can transform the target domain image to a new representation ($ \hat{{X^t}} = \mathcal{F_{PDT}}(X^t)$) to reduce the domain gap while retaining discriminative information. This new representation ($\hat{{X^t}}$) can be used together with a pre-trained FR model to achieve the \textit{HFR} task.

To accomplish this task, we propose to prepend a small network module called ``\proposed'' (PDT) to a pre-trained FR model. A schematic diagram of the proposed framework is depicted in Fig. \ref{fig:framework}. Essentially, we apply this module as a transformation to the target modality images, which generates a \textit{transformed} ($\mathcal{F_{PDT}}(X^t)$) image. 

This can also be viewed as an extension of the DSU \cite{de2018heterogeneous} approach. Instead of adapting the lower layers as in DSU, we prepend a neural network block in front of a pre-trained FR model to handle domain-specific features. However, one of the main restrictions of DSU is that the local features that could be modified are restricted by the design of the face recognition (FR) network. One can only try to adapt or freeze layers already present in the FR network. It should also be noted that the architecture of the low-level layers in the FR network is designed for optimizing the performance in visible spectrum face recognition, and as such the lower layers (and their architecture) may not be optimal for the \textit{HFR} task. This creates a bottleneck that limits the possibility of learning. In \proposed there is more flexibility by changing the architecture of the \sdt block, one can even change the local receptive field by changing the architecture of the \sdt block. One could even perform a neural architecture search \cite{tan2019mnasnet} to optimize the architecture of the PDT block for a specific heterogeneous use case. The architecture we use is more general which fits a larger set of heterogeneous tasks. In this sense, \sdt is more flexible compared to DSU. Moreover, in PDT, the transformation is performed in the pixel space itself, meaning this approach can be applied to several FR architectures as a plug-in module. We have reimplemented DSU  heterogeneous face recognition approach with the recent Iresnet100 pre-trained model as an additional baseline (DSU-Iresnet100).

The \textit{transformed} image can then be passed to a pre-trained FR model to get the embeddings for the \textit{HFR} task. With the proposed approach, we can express the \textit{HFR} problem in the following way: \\
\begin{equation}
  P(Y|X_t, \hat{\Theta}) = P(Y|X_t, [\theta_{PDT}, \Theta_{FR}])
\end{equation}

The parameters of \sdt block ($\theta_{PDT}$) can be learned in a supervised setting using back-propagation. In the forward pass for a tuple ($X^s, X^t$), the $X^s$ image directly passes through the shared pre-trained FR network to produce the embedding. The target image ($X^t$) first passes through the \sdt module ($ \hat{{X^t}} = \mathcal{F_{PDT}}(X^t)$), and then the \textit{transformed} image passes through the shared pre-trained FR model to generate the embedding. Contrastive loss \cite{hadsell2006dimensionality} is employed as the loss function in the training phase, to reduce the distance between these cross-modal embeddings when the identities are the same and increase the distance when the identities are different. The Contrastive loss is given as:

\begin{equation}
\begin{split}
\mathcal{L}_{Contrastive}(\Theta, Y_p, X_s, X_t)= & (1-Y_p)\frac{1}{2}D_W^{2}  \\
            & + Y_p\frac{1}{2}{max(0, m-D_W)}^{2}
\end{split}
\end{equation}

Where $\Theta$ denotes the weights of the network, $X_s, X_t$ denote the heterogeneous pairs and $Y_p$ the label of the pair, i.e., whether they belong to the same identity or not, $m$ is the margin, and $D_W$ is the distance function between the embeddings of the two samples. The label $Y_p=0$, when the identities of subjects in $X_s$ and $X_t$ are the same, and $Y_p=1$ otherwise. The distance function $D_W$ can be computed as the Euclidean distance between the features extracted by the network.

The parameters of the shared FR model are kept frozen during the training and only the parameters of the \sdt module are updated in the backward pass. At the end of the training, the model corresponding to minimum validation loss is selected which is used for the evaluations.

\subsection{Architecture of the \proposed (PDT) block}
%--------------------------------------------

The \proposed block is parameter-efficient and generic, allowing it to be applied to a wide range of \textit{HFR} scenarios. The input and output of the \sdt block are `three-channel' images with the same size. This makes it easy to visualize the output of the proposed \sdt module and to pass the modified images on to pre-trained FR models during inference. This module can also be readily ``plugged in'' to any pre-trained FR pipeline to convert it to an \textit{HFR} pipeline.

\begin{figure}[h]
  \centering
  \includegraphics[width=0.99\linewidth]{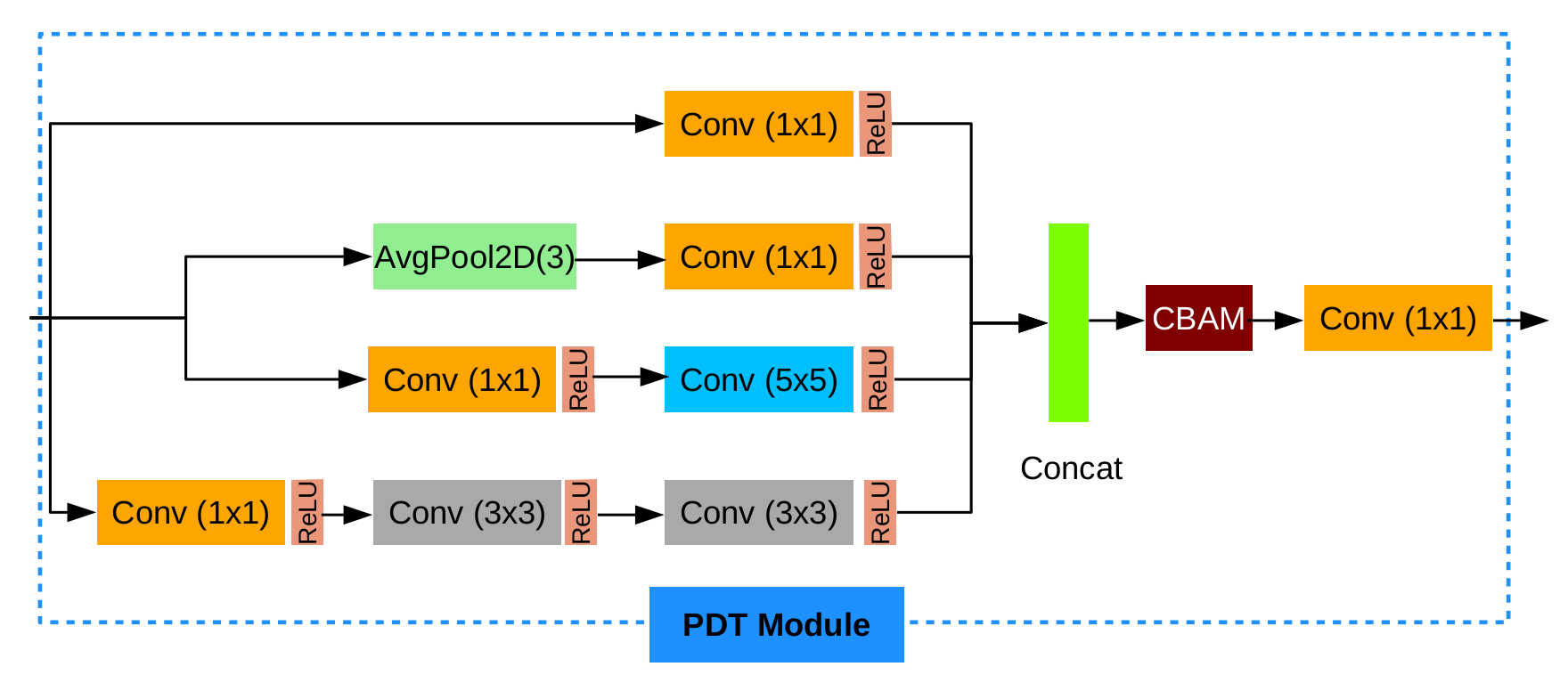}
  \caption{Architecture of the \proposed (PDT) block.}
  \label{fig:sdt_block}
\end{figure}

% can be shortened more

The architecture of the proposed \sdt module is shown in Fig. \ref{fig:sdt_block}. The initial part of the \sdt bock utilizes the design principles inspired from the inception architecture \cite{szegedy2015going}. We follow the idea of multi-scale processing by using different parallel branches with different kernel sizes. Parallel branches were necessary since the receptive field required for various heterogeneous settings differs, and having multi-scale features at the input level aids in a generic design with minimal computational complexity. There are four parallel paths from the input image, 1) a $1 \times 1$ filter, a $3 \times 3$ branch with two sequential filters, a $5 \times 5$ branch, and an average pooling branch. In each of these branches, $1 \times 1$ convolutions are used to reduce the number of output channels. We use ReLU activation after each of the convolution operations. A feature map is formed by concatenating the outputs from each of these branches, which consist of features obtained using filters with different receptive fields. The addition of an attention mechanism helps the network in deciding ``what'' and ``where'' to focus. A Convolutional Block Attention Module (CBAM) \cite{woo2018cbam} attention module was added which achieves this in a simple and parameter efficient way. The CBAM block acts on a feature map along the channel as well as the spatial dimension in a sequential manner. The attention maps obtained are multiplied by the input feature map.  The addition of the CBAM module helps in focusing on meaningful features along the channel and spatial dimensions making the proposed architecture robust to a wide variety of \textit{HFR} scenarios. After the CBAM block, the channel dimension of the output feature map is still high and a $1 \times 1$ convolutional layer is added to reduce the channel dimension to three.

Overall the number of parameters to learn is merely 1.4K. The minimal design enables the network to focus on important features with a minimal parameter overhead. It is to be noted that, this module can be further optimized for specific heterogeneous scenarios. 

\subsection{Pre-trained FR backbone}
%-------------------------------

As described earlier, the \sdt module can be prepended to any pre-trained FR model. Though we perform most of the experiments with \textit{Iresnet100} model \cite{insightface}, this can be extended to many publicly available pre-trained FR models. For the sake of reproducibility, we used publicly available pre-trained face recognition models available from \cite{wang2021facex} \footnote{\url{https://github.com/JDAI-CV/FaceX-Zoo}}. These models were trained on  MS-Celeb-1M-v1c \footnote{\url{http://trillionpairs.deepglint.com/data}} with 72,778 identities and about 3.28M images. In most cases, the pre-trained FR model accepts three-channel images with a resolution of $112 \times 112$. Faces are first aligned and cropped ensuring eye center coordinates fall on pre-fixed points. In the case of single-channel inputs (such as NIR, thermal, etc.), we replicate the same channel to three channels without making any changes to the network architecture. This was necessary since the pre-trained networks were designed to accept three-channel RGB images, and we didn't want to alter the layers/weights of the FR network that would change the performance of the pre-trained network in the RGB images in any manner.

\subsection{Implementation details}
%----------------------------------

The proposed framework is trained with Contrastive loss in a standard Siamese network setting \cite{hadsell2006dimensionality}. The margin parameter is set as 2.0 in all the experiments. We used Adam Optimizer with a learning rate of $0.001$ and trained the model for 20 epochs with a batch size of 90. The framework was implemented in PyTorch using the Bob library \cite{bob2017}. Except for the new \sdt module added to the target channel branch, the whole pre-trained FR model is shared between source and target modalities in the Siamese network. Only the parameters of the \sdt module are adapted during training, while the weights of the FR model remain frozen. The experiments are reproducible and the source code and the protocols will be made available publicly \footnote{\url{https://gitlab.idiap.ch/bob/bob.paper.tifs2022_hfr_prepended_domain_transformer}}.

The proposed method may be applied to a variety of \textit{HFR} scenarios, including VIS-Thermal, VIS-SWIR, and VIS-Low resolution VIS, and so on. Furthermore, components of the proposed framework and the training routine are kept simple to demonstrate the efficacy of the proposed approach.

\section{Experiments}
\label{sec:experiments}
%============================================================================================================================================================

An extensive set of experiments and ablation studies with the proposed approach are presented in this section. Primarily, we have evaluated VIS-Thermal \textit{HFR} performance in four different datasets. We also compare the performance of the proposed approach against other heterogeneous settings such as VIS-NIR, VIS-SWIR, and so on. We further perform several studies evaluating the amount of data required for training the models and also the performance with different FR architectures. 
\subsection{Databases and Protocols}
%------------------------------------

The datasets used in the evaluations are described in the following section.

\textbf{Polathermal dataset:} Pola Thermal dataset \cite{hu2016polarimetric} --Polarimetric and Thermal Database is an \textit{HFR}
dataset collected by the U.S. Army Research Laboratory (ARL). The dataset contains polarimetric LWIR (long-wave infrared) imagery together with color images collected synchronously for 60 subjects. The dataset contains thermal imagery collected for conventional thermal images as well as polarimetric images. In this work, we use the conventional thermal images for our experiments using the reproducible protocols introduced in \cite{de2018heterogeneous}. We follow the same five-fold partitions in which 60 subjects were split into a training set with 25 identities and 35 identities for testing. To compare different methods, the average Rank-1 identification rate is reported from the evaluation set of the five folds.

\textbf{Tufts face dataset:} The Tufts Face Database \cite{panetta2018comprehensive} provides face images captured with different modalities for the \textit{HFR} task. Specifically, we use the thermal images provided in the dataset to evaluate VIS-Thermal \textit{HFR} performance. Overall, there are a total of 113 identities comprising of 39 males and 74 females from different demographic regions. For each subject, images from different modalities are available. For comparison purposes, we follow the procedure followed by authors in \cite{fu2021dvg}, 50 identities are randomly selected from the data as the training set and the remaining subjects were used as the test set. We report the Rank-1 accuracies and Verification rates at false acceptance rates (FAR) 1\% as well as 0.1\% for comparison.

\textbf{ARL-VTF dataset:} In \cite{poster2021large}, authors made available the  DEVCOM Army Research Laboratory Visible-Thermal Face Dataset (ARL-VTF). The dataset contains heterogeneous data from 395 subjects with three visible spectra as well as one thermal (long-wave infrared- LWIR) camera, with over 500,000 images altogether. The dataset contains variability in terms of expressions, pose, and eyewear. We evaluate the models with the protocols originally provided with the dataset. The dataset also provides annotations for face landmarks. Several protocols evaluating the effects of the pose, expressions, and eyewear are also provided with the dataset. The $test$ set for each setting is fixed to enable direct comparisons with state-of-the-art methods. The naming of each protocol is as follows:  Gallery and Probe protocols are designated ``\textbf{G}" and ``\textbf{P}" respectively. ``\textbf{V}" and ``\textbf{T}" denote the visible and thermal images. Categories such as  ``\textbf{B}", ``\textbf{E}", and ``\textbf{P}" denote baseline, expression, and pose sequences. Presence of the ``$*$" symbol indicates any or all sequence categories in the protocol. A subject who does not possess glasses has the tag \textbf{0}, and \textbf{-} and \textbf{+} tags are present for subjects who have their glasses removed or worn respectively.

In summary, the structure of protocol names are as follows:
\setlength{\belowdisplayskip}{0pt} \setlength{\belowdisplayshortskip}{0pt}
\setlength{\abovedisplayskip}{0pt} \setlength{\abovedisplayshortskip}{0pt}

\begin{bnf*}
\bnfprod{set} {``\textbf{G}" \bnfor ``\textbf{P}"} ; \\
\bnfprod{modality} {``\textbf{V}" \bnfor ``\textbf{T}"} ; \\
\bnfprod{sequence} {``\textbf{B}" \bnfor ``\textbf{E}" \bnfor ``\textbf{P}" \bnfor ``*" ; } \\
\bnfprod{eyewear} {``\textbf{0}" \bnfor ``\textbf{-}" \bnfor ``\textbf{+}" ; } \\
\bnfprod{protocol} {\bnfpn{set}, ``\_", \bnfpn{modality}, } \\
\bnfmore{\bnfpn{sequence}, [\bnfpn{eyewear}+] ;}
\end{bnf*}

A detailed description of these protocols can be found in \cite{poster2021large}.

\textbf{CASIA NIR-VIS 2.0 dataset:} The CASIA NIR-VIS 2.0 Face Database \cite{li2013casia} provides images of subjects captured with both visible spectrum as well as near-infrared lighting, with a total of 725 identities. Each subject in the dataset has 1-22 visible images and 5-50 near-infrared (NIR) images. The experimental protocols provided uses a 10-fold cross-validation protocol with 360 identities used for training. The gallery and probe set for evaluation consist of 358 identities. The train and test sets are made with disjoint identities. We perform experiments in each fold and the mean and standard deviation of the performance metrics are reported.

\textbf{SCFace dataset:} The SCFace \cite{grgic2011scface}, dataset contains high quality mugshot for enrollment for FR. The probe samples correspond to surveillance scenarios coming from different cameras and are of low quality. Depending on the distance and quality of probe samples, four different protocols are present. They are close, medium, combined, and far. The ``far'' protocol is the most challenging one.  The dataset contains 4,160 static images (in the visible and infrared spectrum) from 130 subjects.

\textbf{MCXFace Dataset:} We present a new \textit{HFR} dataset named Multi-Channel Heterogeneous Face Recognition dataset (MCXFace). The dataset is derived from 
the HQ-WMCA dataset we have created earlier \cite{heusch2020deep,mostaani2020idiap}. The dataset contains images of 51 subjects collected in different channels under three different sessions and various illumination conditions. The channels available are color (RGB), thermal, near-infrared (850 nm), short-wave infrared (1300 nm), Depth, Stereo depth, and depth estimated from RGB images using 3DDFA \cite{guo2020towards} method. All the channels are registered spatially and temporally across all the modalities. The details about the sensors and data collection sessions can be found in our earlier work \cite{heusch2020deep,mostaani2020idiap}. In the MCXFace dataset, only bonafide samples are present. Further, the files are divided into $train$ and $dev$ sets with a disjoint set of identities to make experiments in different homogeneous and heterogeneous settings possible. For each of the protocols, we have created five different folds, by randomly dividing the subjects in $train$ and $dev$ partitions. Each of the protocol names is of the following form: $<SOURCE>-<TARGET>-split<split>$. In addition to the images, annotations for left and right eye centers for all the images are also provided. The dataset will be available publicly in the following link \footnote{\url{https://www.idiap.ch/dataset/mcxface}}. %Each file is a ``.jpg'' file with a resolution of $1920 \times 1200$.

\textbf{CUFSF dataset:} The CUHK Face Sketch FERET Database (CUFSF) \cite{zhang2011coupled} contains 1194 faces from the FERET dataset \cite{phillips1998feret} and each image in the FERET dataset has a corresponding sketch image drawn by an artist. The dataset is challenging since the sketches have more shape exaggerations compared to the source photos. We follow the same protocol as reported in \cite{fang2020identity}, where 250 identities were used for training the model, and the rest of 944 identities are reserved as the testing set. The Rank-1 accuracies are reported for comparison.

\subsection{Metrics}
%--------------------

To evaluate the models we follow several different metrics corresponding to previous literature. We have used a subset of metrics from the following performance metrics, Area Under the Curve (AUC), Equal Error Rate (EER), Rank-1 identification rate, Verification Rate with different false acceptance rates (0.01\%, 0.1\%, 1\%, and 5\%).

\subsection{Experimental results}
%----------------------------------------------

The experiments performed in the different datasets and the results are discussed in this section.

\subsubsection{\textbf{Experiments with Polathermal dataset}}
%%%----------------------------------------------

Here we perform experiments in thermal to visible recognition scenarios.
The results in Table. \ref{tab:polathermal} shows the average Rank-1 identification rate in the five protocols of the Polathermal `thermal to visible protocols' (using the reproducible protocols in \cite{de2018heterogeneous}). The reimplemented DSU-Iresnet100 baseline achieves better results compared to the original model from \cite{de2018heterogeneous}, indicating that the use of a better pre-trained model improves the results. It can be seen that the proposed approach achieves an average Rank-1 accuracy of 97.1\% with a standard deviation of 1.3\%, which is much greater than the results from other baselines reported in the literature.

\begin{table}[ht]
\caption{Pola Thermal - Average rank one recognition rate}
\label{tab:polathermal}
\begin{center}
  \begin{tabular}{lrr}
    \toprule
    \textbf{Method} & \textbf{Mean (Std. Dev.)} & \textbf{Info} \\ \midrule
    
    DPM in \cite{hu2016polarimetric}   & 75.31 \% (-)     & Paper \\ 
    CpNN in \cite{hu2016polarimetric}  & 78.72 \% (-)     & Base-  \\ 
    PLS in \cite{hu2016polarimetric}   & 53.05\% (-)      & ines    \\  \hline

    LBPs + DoG features in \cite{liao2009heterogeneous} & 36.8\% (3.5)      & Repro-    \\ 
    ISV in \cite{de2016heterogeneous}       & 23.5\% (1.1)      & ducible   \\ 
    GFK in \cite{sequeira2017cross}             & 34.1\% (2.9)      & baselines \\

    DSU(Best Result) \cite{de2018heterogeneous} & 76.3\% (2.1) & Reproducible \\
    
    DSU-Iresnet100 & 88.2\% (5.8) & Reproducible \\
    \hline
    \rowcolor{Gray}
    \textbf{\sdt (Proposed)}  & \textbf{97.1\% (1.3)} & Reproducible\\
    \bottomrule 
  \end{tabular}
\end{center}
\end{table}

\subsubsection{\textbf{Experiments with Tufts face datasets}}
%%%----------------------------------------------

\begin{figure}[h]
  \centering
  \includegraphics[width=0.99\linewidth]{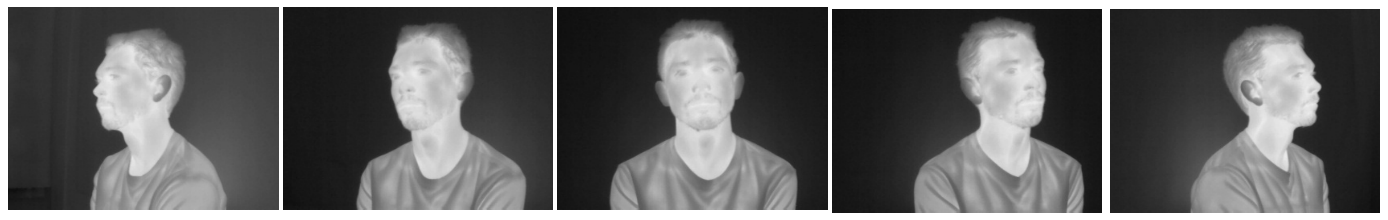}
  \caption{Challenging pose variations in VIS-Thermal protocol of the TUFTS face dataset.}
  \label{fig:tuft_pose}
\end{figure}

Table. \ref{tab:tufts} shows the performance of the proposed approach against other state-of-the-art methods in the VIS-Thermal protocol of the Tufts face dataset. The Tufts face dataset is very challenging due to pose and other types of variations. The challenging pose variations in Tufts face dataset are depicted in Fig. \ref{fig:tuft_pose}. Performance of even visible spectrum face recognition systems degrade in such extreme yaw angles, so it is expected that the performance of \textit{HFR} would also degrade. Nevertheless, it can be seen that the proposed approach achieves the best results in verification rate, and is only second to DVG-Face \cite{fu2021dvg} in Rank-1 accuracy, showing the effectiveness of the proposed approach.

\begin{table}[h]
  \centering
  \caption{Experimental results on VIS-Thermal protocol of the Tufts Face dataset.}
  \label{tab:tufts}
  % \resizebox{0.45\textwidth}{!}{
  \begin{tabular}{lccc}
    \toprule
      Method & Rank-1 & VR@FAR=1$\%$ & VR@FAR=0.1$\%$  \\ \midrule
      LightCNN \cite{Wu2018ALC} & 29.4 & 23.0 & 5.3 \\
      DVG \cite{fu2019dual} & 56.1 & 44.3 & 17.1 \\
      DVG-Face \cite{fu2021dvg} & \textbf{75.7} & 68.5 & 36.5 \\ 
      DSU-Iresnet100 & 49.7 & 49.8 & 28.3 \\ \midrule  \rowcolor{Gray}
      
      \textbf{\sdt (Proposed)} & 65.71 & \textbf{69.39} & \textbf{45.45} \\ \bottomrule
  % }
  \end{tabular}
\end{table}

\subsubsection{\textbf{Experiments with CASIA-VIS-NIR 2.0 dataset}}
%%%----------------------------------------------

Though most of the focus is given to Thermal-VIS \textit{HFR}, we perform experiments in the CASIA-VIS-NIR 2.0
dataset to showcase the effectiveness of the proposed approach in other heterogeneous scenarios, specifically NIR-VIS recognition.
As it can be seen from the baselines, the domain gap appears to be less in this scenario and even some of the pre-trained FR model trained using VIS modality achieves reasonable performance. Due to this, the evaluations are done with even tighter thresholds and VR@FAR=0.1$\%$  and VR@FAR=0.01$\%$ are used for comparison. There are 10 sub-protocols in the dataset, and we report the mean and standard deviation of all ten folds in the comparison. The results are
compared in Tab. \ref{tab:casia}. From the results, it can be seen that the proposed approach achieves the best performance compared to other state-of-the-art methods. The superior performance indicates that our framework generalizes across several heterogeneous scenarios.

\begin{table}[h]
  \centering
  \caption{Experimental results on CASIA NIR-VIS 2.0.}
  \label{tab:casia}
  \resizebox{0.47\textwidth}{!}{
  \begin{tabular}{l|ccc}
    \toprule
      Method & Rank-1 & VR@FAR=0.1$\%$ & VR@FAR=0.01$\%$ \\
      \midrule
      IDNet \cite{Reale2016SeeingTF} & 87.1$\pm$0.9 & 74.5 & - \\
      HFR-CNN \cite{saxena2016heterogeneous} & 85.9$\pm$0.9 & 78.0 & - \\
      Hallucination \cite{Lezama2017NotAO} & 89.6$\pm$0.9 & - & - \\
      % DLFace \cite{peng2019dlface} & 98.68 & - & - \\
      TRIVET \cite{XXLiu:2016} & 95.7$\pm$0.5 & 91.0$\pm$1.3 & 74.5$\pm$0.7 \\
      W-CNN \cite{DBLP:journals/corr/abs-1708-02412} & 98.7$\pm$0.3 & 98.4$\pm$0.4 & 94.3$\pm$0.4 \\
      PACH \cite{duan2019pose} & 98.9$\pm$0.2 & 98.3$\pm$0.2 & - \\
      RCN \cite{deng2019residual} & 99.3$\pm$0.2 & 98.7$\pm$0.2 & - \\
      MC-CNN \cite{8624555} & 99.4$\pm$0.1 & 99.3$\pm$0.1 & - \\
      DVR \cite{XWu:2019}  & 99.7$\pm$0.1 & 99.6$\pm$0.3 & 98.6$\pm$0.3 \\
      
      DVG \cite{fu2019dual} & 99.8$\pm$0.1 & 99.8$\pm$0.1 & 98.8$\pm$0.2 \\
      DVG-Face \cite{fu2021dvg} & 99.9$\pm$0.1 & 99.9$\pm$0.0 & 99.2$\pm$0.1 \\ \hline
      %\textbf{Proposed} & \textbf{99.97}$\pm$0.01 & \textbf{99.95}$\pm$0.02 & \textbf{99.81}$\pm$0.04 \\ on dev todo on eval (trained only with train set 0.2% as val)
      \rowcolor{Gray}
      \textbf{\sdt (Proposed)} & \textbf{99.95}$\pm$0.04 & \textbf{99.94}$\pm$0.03 & \textbf{99.77}$\pm$0.09 \\ \bottomrule % on eval
  \end{tabular}}
\end{table}

\subsubsection{\textbf{Experiments with SCFace dataset}}
%%%----------------------------------------------

We have performed a set of experiments on the SCFace dataset using the protocols provided for visible images in the dataset. In this dataset, the heterogeneity arises from the quality difference between gallery and probe images, i.e., gallery images are high-resolution mugshots and probe images and low-resolution images coming from a surveillance camera. The results are tabulated in Table. \ref{tab:scface}, the values reported corresponds to the ``evaluation'' set of the protocols. The baseline model is a pre-trained \textit{Iresnet100} model as used in other experiments, and the rows with \sdt denotes the proposed model where the \sdt model is trained using contrastive training. Results with reimplemented DSU-Iresnet100 model is also added for comparison. It can be seen that with the proposed approach the performance of the baseline model improves. The improvement is obvious in the ``far'' protocol where the quality of the probe images is very poor. It can be seen the \sdt module, in this case, helps in learning quality and resolution invariant features improving the results.

\begin{table}[h]
  \caption{Performance of the proposed approach in the SCFace dataset, the Baseline is a pretrained \textit{Iresnet100} model, and the \sdt is  with the proposed approach. }
  \label{tab:scface}
  \centering
  \resizebox{\columnwidth}{!}{%
  \begin{tabular}{lcrrrr}
  \toprule
  Protocol             & Method & AUC   & EER   & Rank-1    & \begin{tabular}[c]{@{}c@{}} VR@\\FAR=0.1\% \end{tabular} \\ \midrule
  \multirow{3}{*}{Close} & Baseline & 100.0   & 0.00    & 100.0   & 100.0                \\
                        &DSU-Iresnet100 & 100.0 & 0.00 & 100.0 & 100.0    \\
                                  % \rowcolor{Gray}

                             & \cellcolor{Gray} \textbf{\sdt}  &\cellcolor{Gray}  100.0   &\cellcolor{Gray}  0.00    &\cellcolor{Gray}  100.0   &\cellcolor{Gray}  100.0                    \\ \midrule

\multirow{3}{*}{Medium}   & Baseline & 99.81 & 2.33 & 98.60  & 92.09              \\
&DSU-Iresnet100 & 99.95 & 1.39 & 98.98 & 93.25 \\
                                % \rowcolor{Gray}

                             &\cellcolor{Gray}  \textbf{\sdt} & \cellcolor{Gray} 99.96 &\cellcolor{Gray}  0.93 &\cellcolor{Gray}  99.07 &\cellcolor{Gray} 95.81                 \\ \midrule
\multirow{3}{*}{Combined}    & Baseline & 98.59 & 6.67 & 91.01 & 77.67          \\
                                % \rowcolor{Gray}
                                &DSU-Iresnet100 & 98.91 & 4.96 &92.71 & 80.93  \\
                             &\cellcolor{Gray}  \textbf{\sdt} &\cellcolor{Gray}  99.06 &\cellcolor{Gray}  4.50  &\cellcolor{Gray}  93.18 & \cellcolor{Gray} 82.02           \\      \midrule
\multirow{3}{*}{Far}     & Baseline & 96.59 & 9.37 & 74.42 & 49.77           \\
                                % \rowcolor{Gray}
                                &DSU-Iresnet100 & 97.18 & 8.37 & 79.53 & 58.26  \\

                             & \cellcolor{Gray} \textbf{\sdt}   &\cellcolor{Gray}  98.31 &\cellcolor{Gray}  6.98 &\cellcolor{Gray}  84.19 &\cellcolor{Gray} 60.00            \\   

  \bottomrule
  \end{tabular}
  }
  \end{table}

\subsubsection{\textbf{Experiments with MCXFace dataset}}
%%%----------------------------------------------

The experiments in the MCXFace dataset gives a unique opportunity to evaluate the performance of the models in several heterogeneous scenarios, including VIS-Thermal, 
VIS-Depth, VIS-SWIR, VIS-NIR, and so on. 
The results containing the average performance among five folds are shown in Table. \ref{tab:mcxface}. For each modality, (eg. VIS-Thermal), the values reported are aggregated over the five folds in the dataset. In each protocol, we perform a baseline evaluation which is essentially using the pre-trained \textit{Iresnet100} FR model on the new modality. The lower baseline performance indicates a large domain gap.  For example, running a vanilla FR model itself achieves excellent performance for NIR and SWIR channels, whereas the performance is poor for depth and thermal channels. In the Table. \ref{tab:mcxface}, the rows with \sdt show the results with our proposed approach. In addition, the reimplemented DSU-Iresnet100 model is also added as a baseline.  It can be seen that the performance improves greatly in the case of the thermal channel. However, for the depth channel, even though the performance improves, the final performance is not satisfactory, it indicates that the depth channel needs a different treatment, and using the depth data as range images may not be the optimal choice. Using representations like normals or point clouds could be well suited for the depth modality.

\begin{table}[h]
  \caption{Performance of the proposed approach in the MCXFace dataset, the Baseline is a pre-trained \textit{Iresnet100} model, and the \sdt is with the proposed approach. }
  \label{tab:mcxface}
  \centering
  \resizebox{\columnwidth}{!}{%
  \begin{tabular}{lcrrrr}
  \toprule
  Protocol             & Method & AUC   & EER   & Rank-1    & \begin{tabular}[c]{@{}c@{}} VR@\\FAR=0.1\% \end{tabular} \\ \midrule
  \multirow{3}{*}{VIS-Thermal} & Baseline & 84.45 $\pm$ 3.70  & 22.07 $\pm$ 2.81 & 47.23 $\pm$ 3.93  & 19.76 $\pm$ 2.73   \\
  & DSU-Iresnet100 & 98.12 $\pm$ 0.75 & 6.58 $\pm$ 1.35 & 83.43 $\pm$ 5.47 & 52.32 $\pm$ 10.06 \\

                             & \cellcolor{Gray} \textbf{\sdt}      & \cellcolor{Gray}98.43 $\pm$ 0.78  & \cellcolor{Gray} 6.52 $\pm$ 1.45  & \cellcolor{Gray} 84.52 $\pm$ 5.36  &\cellcolor{Gray} 59.05 $\pm$ 13.95  \\ \midrule

\multirow{3}{*}{VIS-Depth}   & Baseline & 53.33 $\pm$ 4.20  & 48.11 $\pm$ 3.40 & 5.19 $\pm$ 1.20   & 0.00 $\pm$ 0.00      \\
                                % \rowcolor{Gray}
& DSU-Iresnet100 & 52.99 $\pm$ 4.74  & 48.37 $\pm$ 0.16 & 4.91 $\pm$ 3.40 & 0.45 $\pm$ 0.62 \\
                             & \cellcolor{Gray} \textbf{\sdt}      & \cellcolor{Gray} 62.16 $\pm$ 5.41  & \cellcolor{Gray} 41.71 $\pm$ 4.31 & \cellcolor{Gray} 9.11 $\pm$ 3.07   & \cellcolor{Gray}0.70 $\pm$ 1.05     \\ \midrule
\multirow{3}{*}{VIS-SWIR}    & Baseline & 100.00 $\pm$ 0.00 & 0.03 $\pm$ 0.02  & 100.00 $\pm$ 0.00 & 99.95 $\pm$ 0.10   \\
                                % \rowcolor{Gray}
& DSU-Iresnet100 & 99.99 $\pm$ 0.01 & 0.16 $\pm$ 0.20 & 99.90 $\pm$ 0.21 & 99.65 $\pm$ 0.39 \\
                             & \cellcolor{Gray} \textbf{\sdt}      &\cellcolor{Gray}  100.00 $\pm$ 0.00 &\cellcolor{Gray}  0.06 $\pm$ 0.08  &\cellcolor{Gray}  99.95 $\pm$ 0.12  &\cellcolor{Gray}  99.85 $\pm$ 0.23   \\ \midrule
\multirow{3}{*}{VIS-NIR}     & Baseline & 100.00 $\pm$ 0.00 & 0.00 $\pm$ 0.00  & 100.00 $\pm$ 0.00 & 100.00 $\pm$ 0.00  \\
                                % \rowcolor{Gray}
& DSU-Iresnet100 & 100.00 $\pm$ 0.00 & 0.00 $\pm$ 0.00 & 100.00 $\pm$ 0.00 & 100.00 $\pm$ 0.00 \\
                             & \cellcolor{Gray} \textbf{\sdt}      & \cellcolor{Gray} 100.00 $\pm$ 0.00 &\cellcolor{Gray}  0.00 $\pm$ 0.00  &\cellcolor{Gray}  100.00 $\pm$ 0.00 &\cellcolor{Gray}  100.00 $\pm$ 0.00  \\ 

  \bottomrule
  \end{tabular}
  }
  \end{table}

\subsubsection{\textbf{Experiments with ARL-VTF dataset}}
%%%----------------------------------------------

The ARL-VTF offers a large-scale VIS-Thermal dataset with different variations such as pose, expressions, and eyewear making it possible to perform experiments to evaluate the effect of different factors. We have used the standard protocols shipped with the datasets for our evaluations. The test sets of different protocols are fixed and the best model selected from the cross-validation was used to evaluate the performance. The performance of the proposed approach against
the state-of-the-art methods is shown in Table. \ref{tab:comparison_table_arl}. It can be seen that the proposed approach achieves state-of-the-art performance in
most cases. The performance improvement is very clear in challenging protocols with pose variations (P\_TP-). In the $P\_TB0-G\_VB0-$ protocol, where all the images are frontal, the proposed method achieves a verification rate of 98.57\% for FAR1\% (and a corresponding Rank-1 accuracy of 99.14\%), whereas in the $P\_TP0-G\_VB0-$ protocol where the samples contain pose variations the verification rate drops to 60.80\% ( and a corresponding Rank-1 accuracy of 60.23\%). The Rank-1 accuracy for just frontal faces is 99.14\% whereas with pose variations it is 60.23\%. It can be seen that this drop is present in all other methods as well, indicating that \textit{HFR} also suffers performance degradation with pose variations as with homogeneous face recognition.

\begin{table*}[h]
  \caption{Verification performance comparisons with state-of-the-art methods for different protocols in ARL-VTF dataset.}
  \centering
  \resizebox{2\columnwidth}{!}{%
  \begin{tabular}{rrrrrrrrrrr}%LLLLLLLLLL lllllllllll
  \toprule 
  \multicolumn{1}{c}{} & \multicolumn{1}{l}{} & \multicolumn{4}{c}{Gallery G\_VB0-} & \multicolumn{4}{c}{Gallery G\_VB0+} \\
  \cmidrule(lr){3-6} \cmidrule(lr){7-10}
  \multicolumn{1}{l}{Probes} & \multicolumn{1}{l}{Method} & \nm{AUC} & \nm{EER} &  \nm{VR@FAR=1\%} & \nm{VR@FAR=5\%} & \nm{AUC} & \nm{EER} &  \nm{VR@FAR=1\%} & \nm{VR@FAR=5\%} \\
  
  \midrule  {\multirow{4}{*}{P\_TB0}}
  & \nm{Raw} & 61.37 & 43.36 & 3.13  & 11.28 & 62.83 & 42.37 & 4.19 & 13.29 \\
  & \nm{Pix2Pix \cite{isola2017image}} & 71.12 & 33.80  &  6.95 & 21.28 & 75.22 & 30.42 & 8.28 & 27.63 \\
  & \nm{GANVFS} \cite{Zhang2017} & 97.94 & 8.14 & 75.00  & 88.93 & 98.58 & 6.94 & 79.09 & 91.04 \\
  & \nm{Di \etal \cite{di2019polarimetric}} & 99.28 & 3.97 & 87.95  & 96.66 & 99.49 & 3.38 & 90.52 & 97.81 \\
  & \nm{Fondje \etal \cite{fondje2020cross}} & 99.76 & 2.30 & 96.84 & 98.43 & 99.87 & 1.84 & 97.29 & 98.80 \\
  \cmidrule(lr){1-2}
  \rowcolor{Gray}
  & \nm{\textbf{\sdt (Proposed)}} &\textbf{99.95}  &\textbf{1.13}  &\textbf{98.57}  &\textbf{100.00} &\textbf{99.95}  &\textbf{1.14}  &\textbf{98.57}  &\textbf{100.00}  \\
  \cmidrule(lr){1-10}
  
  {\multirow{4}{*}{P\_TB-}}
  & \nm{Raw} & 61.14 & 41.64 & 2.77  & 16.11 & 57.61 & 44.73 & 1.38 & 6.11 \\
  & \nm{Pix2Pix} \cite{isola2017image} & 68.77 & 38.02  &  6.69 & 20.28 & 52.11 & 48.88 & 2.22 & 4.66 \\
  & \nm{GANVFS} \cite{Zhang2017} & 99.36 & 3.77 & 84.88  & 97.66 & 87.34 & 18.66 & 7,00 & 29.66 \\
  & \nm{Di \etal \cite{di2019polarimetric}} & 99.63 & 2.66 & 91.55  & 98.88 & 89.24 & 19.49 & 16.33 & 41.22 \\
  & \nm{Fondje \etal \cite{fondje2020cross}} & 99.83 & 1.95 & 96.00 & 99.48 & 99.03 & 4.79 & 85.56 & 95.86 \\
  \cmidrule(lr){1-2}
  \rowcolor{Gray}
  & \nm{\textbf{\sdt (Proposed)}} &\textbf{99.96}  &\textbf{1.18}  &\textbf{98.67}  &\textbf{100.00} &\textbf{99.94}  &\textbf{1.33}  &\textbf{98.67}  &\textbf{100.00}  \\
  \cmidrule(lr){1-10}
  
  {\multirow{4}{*}{P\_TE0}}
  & \nm{Raw} & 61.40 & 41.96 & 3.40  & 12.18 & 62.50 & 41.38 & 4.60 & 13.25 \\
  & \nm{Pix2Pix} \cite{isola2017image} & 69.10 & 35.98  &  7.01 & 16.44 & 73.97 & 31.87 & 7.93 & 19.60 \\
  & \nm{GANVFS} \cite{Zhang2017} & 96.81 & 10.51 & 70.41  & 84.00 & 97.73 & 8.90 & 74.20 & 86.80 \\
  & \nm{Di \etal \cite{di2019polarimetric}} & 98.46 & 6.44 & 81.11  & 92.49 & 98.89 & 5.60 & 84.23 & 93.94 \\
  & \nm{Fondje \etal \cite{fondje2020cross}} & 98.95 & 3.61 & 92.61 & 96.88 & 99.01 & 3.57 & 92.69 & 96.93 \\
    \cmidrule(lr){1-2}
    \rowcolor{Gray}
  & \nm{\textbf{\sdt (Proposed)}} &\textbf{99.90}  &\textbf{1.72}  &\textbf{97.43}  &\textbf{99.77} &\textbf{99.90}  &\textbf{1.72}  &\textbf{97.43}  &\textbf{99.77}  \\
  \cmidrule(lr){1-10}
  
  {\multirow{4}{*}{P\_TE-}}
  & \nm{Raw} & 63.26 & 42.34 & 4.66  & 16.28 & 59.33 & 43.17 & 2.04 & 8.00 \\
  & \nm{Pix2Pix} \cite{isola2017image} & 68.78 & 36.24  &  7.75 & 18.06 & 51.05 & 49.11 & 2.26 & 4.95 \\
  & \nm{GANVFS} \cite{Zhang2017} & 98.66 & 5.93 & 73.17  & 92.82 & 83.68 & 22.41 & 6.77 & 22.13 \\
  & \nm{Di \etal \cite{di2019polarimetric}} & 99.30 & 3.84 & 82.55  & 97.44 & 86.12 & 21.68 & 9.88 & 31.62 \\
  & \nm{Fondje \etal \cite{fondje2020cross}} & 99.83 & 2.27 & 95.66 & 99.48 & 99.48 & 3.05 & 89.45 & 98.07 \\
    \cmidrule(lr){1-2}
    \rowcolor{Gray}
  & \nm{\textbf{\sdt (Proposed)}} &\textbf{99.95}  &\textbf{0.93}  &\textbf{99.07}  &\textbf{100.00} &\textbf{99.90}  &\textbf{1.73}  &\textbf{97.87}  &\textbf{100.00}  \\
  \cmidrule(lr){1-10}
  {\multirow{4}{*}{P\_TP0}}
  & \nm{Raw} & 55.24 & 46.25 & 2.23  & 8.25 & 55.10 & 46.34 & 2.91 & 8.74 \\
  & \nm{Pix2Pix} \cite{isola2017image} & 54.86 & 47.22  &  3.13 & 9.78 & 56.50 & 46.03 & 4.01 & 10.84 \\
  & \nm{GANVFS} \cite{Zhang2017} & 63.70 & 41.66 & 16.55  & 23.73 & 65.58 & 40.19 & 17.95 & 25.68 \\
  & \nm{Di \etal \cite{di2019polarimetric}} & 65.06 & 40.24 & 17.33  & 24.56 & 67.13 & 38.67 & 18.91 & 26.46 \\
  & \nm{Fondje \etal \cite{fondje2020cross}} & 66.26 & 38.05 & 22.18 & 30.72 & 68.39 & 36.86 & 22.64 & 31.81 \\
    \cmidrule(lr){1-2}
    \rowcolor{Gray}
  & \nm{\textbf{\sdt (Proposed)}} &\textbf{87.56}  &\textbf{20.57}  &\textbf{60.80}  &\textbf{68.86} &\textbf{87.51}  &\textbf{20.57}  &\textbf{60.86}  &\textbf{68.86}  \\
  \cmidrule(lr){1-10}
  {\multirow{4}{*}{P\_TP-}}
  & \nm{Raw} & 55.48 & 45.98 & 3.25  & 8.47 & 56.82 & 44.74 & 2.09 & 7.57 \\
  & \nm{Pix2Pix} \cite{isola2017image} & 54.31 & 47.04  &  2.93 & 8.44 & 50.08 & 49.67 & 0.60 & 4.33 \\
  & \nm{GANVFS} \cite{Zhang2017} & 65.79 & 40.35 & 17.84  & 25.48 & 59.51 & 44.04 & 4.29 & 15.47 \\
  & \nm{Di \etal \cite{di2019polarimetric}} & 67.27 & 39.00 & 18.16  & 26.02 & 60.10 & 43.57 & 5.77 & 15.97 \\
  & \nm{Fondje \etal \cite{fondje2020cross}} & 68.24 & 37.60 & 23.09 & 33.54 & 63.29 & 41.79 & 18.79 & 27.93 \\
    \cmidrule(lr){1-2}
    \rowcolor{Gray}
  & \nm{\textbf{\sdt (Proposed)}} &\textbf{87.78}  &\textbf{20.40}  &\textbf{65.33}  &\textbf{71.20} &\textbf{87.30}  &\textbf{20.65}  &\textbf{60.00}  &\textbf{69.87}  \\
  \cmidrule(lr){1-10}
  {\multirow{4}{*}{P\_TB+}}
  & \nm{Raw} & 59.52 & 42.60 & 4.66  & 6.00 & 78.26 & 29.77 & 3.88 & 21.33 \\
  & \nm{Pix2Pix} \cite{isola2017image} & 59.68 & 41.72  &  3.33 & 3.33 & 67.08 & 36.44 & 2.68 & 11.11 \\
  & \nm{GANVFS} \cite{Zhang2017} & 87.61 & 20.16 & 20.55  & 44.66 & 96.82 & 8.66 & 46.77 & 83.00 \\
  & \nm{Di \etal \cite{di2019polarimetric}} & 91.11 & 17.43 & 22.33  & 55.66 & 97.96 & 7.21 & 60.11 & 88.70 \\
  & \nm{Fondje \etal \cite{fondje2020cross}}  & 99.28 & 5.32 & 89.21 & 94.79 & \textbf{99.97} & \textbf{0.73} & \textbf{99.47} & \textbf{100.00} \\
    \cmidrule(lr){1-2}
    \rowcolor{Gray}
  & \nm{\textbf{\sdt (Proposed)}} &\textbf{99.48}  &\textbf{4.11}  &\textbf{89.33}  &\textbf{97.33} &99.60  &4.00  &90.00 &97.33  \\
  \bottomrule
  
  \end{tabular}
  }
  \label{tab:comparison_table_arl}
\end{table*}

% Done from bottom until here

\subsubsection{\textbf{Experiments with CUFSF dataset}}
%%%----------------------------------------------

Here we perform experiments with the challenging sketch to photo recognition task. The Rank-1 accuracies obtained with the baseline and the other methods are shown in Table. \ref{tab:cufsf}, using the protocols in \cite{fang2020identity}. It can be seen that the proposed approach obtains a Rank-1 accuracy of 71.08\%. Nevertheless, the absolute accuracy in the sketch to photo recognition is low. Several methods in literature have shown (on a different protocol) that the sketch recognition performance could be improved with specifically designed features \cite{koley2021gammadion} and specially designed neural network models \cite{luo2022memory}. The sketch modality is very different compared to other heterogeneous imaging modalities we have considered so far like thermal, near-infrared, and so on. Though the CUFSF dataset contains viewed hand-drawn sketch images \cite{klum2014facesketchid} which appears to be holistically similar for humans, there exists a  domain gap in the context of an automatic face recognition system as evidenced from the baseline performance. The pretrained model achieves a Rank-1 accuracy of 56.57\%, indicating the domain gap. The modalities like thermal, NIR, and SWIR,  were all the ``imaging'' modalities and it shares the same high-level representation of the face, but a different aspect of the face. In sketch recognition, the sketch images contain exaggerations depending on the artist, and may not optimally preserve the discriminative information which a face recognition network might be looking for. This could explain the larger performance gap of sketch-photo recognition compared to the performance in other imaging modalities.

\begin{table}[ht]
\caption{CUFSF: Rank-1 recognition in sketch to photo recognition}
\label{tab:cufsf}
\begin{center}
  \begin{tabular}{lrr}
    \toprule
    \textbf{Method} & Rank-1 \\ \midrule
    Baseline & 56.57 \\
    \rowcolor{Gray}
    IACycleGAN \cite{fang2020identity} &64.94 \\
    DSU-Iresnet100 & 67.06 \\ \hline
    \rowcolor{Gray} 
    \textbf{\sdt (Proposed)}  & \textbf{71.08} \\
    \bottomrule 
  \end{tabular}
\end{center}
\end{table}

\subsection{Analysis of the Framework}
%----------------------------------------------

To further understand the effectiveness of the proposed approach, we have performed additional experiments on the ARL-VTF dataset due to the large number of subjects present. Specifically, we performed experiments to understand the effect of the amount of training data, the type of supervision used in training, and the performance of different FR architectures. All these experiments were performed on the $G\_VB0-P\_TB-$ protocol of the ARL-VTF dataset.

\subsubsection{\textbf{Performance with limited amount training data}}
%--------------------------------------------------------------------------------------

The amount of paired training data available to train \textit{HFR} model is often limited and expensive to acquire. In this regard, we perform a set of experiments to understand how the amount of training data available to train the model affects its performance. We used the ARL-VTF dataset for this set of experiments because it had a larger number of subjects. The test samples are kept the same for this set of experiments, with the only difference being the amount (or percentage) of training and validation samples. We start with using 100\% of the training data for training the \sdt module and gradually lower the number of samples in the intervals of 10\%, and finally 1\% intervals. For context, we also note the number of subjects in the training set for these scenarios. The results of this set of experiments are tabulated in Table. \ref{tab:ablation_percentage}. Remarkably, the proposed approach achieves a Rank-1 accuracy
of 94.67\% with just 2\% percentage of the training data, for context, just with data from 4 subjects. This could be due to the parameter efficiency of our approach. 
The learnable component of the \sdt block contains approximately just 1.4K parameters, and hence requires a very minimal amount of data to achieve good performance. This is an important
observation, as \textit{HFR} datasets are often small in size.

\begin{table}[h]
  \caption{Experiments with subsets of training data. The test set is kept the same in all experiments.}
  \label{tab:ablation_percentage}
  \centering
  \resizebox{\columnwidth}{!}{%
  \begin{tabular}{@{}crrrrrrr@{}}
  \toprule
  \begin{tabular}[c]{@{}c@{}c@{}}\% of \\ training \\ data\end{tabular} & Subjects & AUC   & EER  & Rank-1& \begin{tabular}[c]{@{}c@{}} VR@\\FAR=0.1\% \end{tabular} \\ \midrule
  
  1\%                                                                      &2         & 83.25 & 25.33 & 20.67 & 5.33   \\
  2\%                                                                      &4         & 99.15 & 5.20  & 94.67 & 85.33  \\
  3\%                                                                      &7         & 98.46 & 3.33  & 93.33 & 88.00     \\
  4\%                                                                      &9         & 98.91 & 3.33  & 93.33 & 85.33  \\
  5\%                                                                      & 11       & 98.55 & 3.33 & 96.67  & 89.33  \\
  10\%                                                                     & 23       & 99.39 & 3.33 & 96.67  & 92.00     \\
  20\%                                                                     & 47       & 99.73 & 3.33 & 97.33  & 96.67  \\
  30\%                                                                     & 70       & 99.77 & 3.33 & 96.00     & 92.67  \\
  40\%                                                                     & 94       & 99.77 & 2.68 & 97.33  & 95.33  \\
  50\%                                                                     & 118      & 99.95 & 1.36 & 99.33  & 96.67  \\
  60\%                                                                     & 141      & 99.9  & 2.67 & 96.67  & 96.67  \\ 
  70\%                                                                     & 165      & 99.68 & 3.33 & 96.67  & 96.00     \\ 
  80\%                                                                     & 188      & 99.67 & 3.33 & 96.67  & 96.00     \\
  90\%                                                                     & 212      & 99.8  & 2.79 & 96.67  & 96.00     \\
  100\%                                                                    & 235      & 99.96 & 1.18 & 99.33  & 96.67  \\ \bottomrule
  \end{tabular}
  }
  \end{table}

\subsubsection{\textbf{Experiments with unpaired images}}
%----------------------------------------------------------------------
So far in the experiments, the networks were trained using contrastive loss in a supervised manner. We assume that paired heterogeneous samples are available at the training time. Here we try to emulate an unpaired setting, meaning we do not have paired heterogeneous samples, and we do not have the identity information of the samples. In this scenario, we supervise our framework to match the feature distributions of source and target modalities using Maximum Mean Discrepancy (MMD) \cite{gretton2012kernel} loss. We have experimented with applying the MMD loss in three different ways, 1) 
the MMD loss was applied to both the \textit{transformed} and source images, thereby attempting to match the FR network's inputs.(ip),  2) The MMD loss is applied to the embeddings obtained from the FR network (op). and 3) where both output embeddings and inputs were supervised by MMD loss (ip+op).  

The results obtained from this set of experiments are shown in Table. \ref{tab:training_arl_mmd}. The first row shows the baseline with a pre-trained \textit{Iresnet100} and the last row shows the results with supervised learning with the \sdt approach. From the results, it can be seen that even in unpaired settings the proposed framework performs reasonably well. The best results are obtained when MMD is used on both the input and output. This indicates that the proposed framework can be employed even if paired samples aren't available. It was possible to achieve reasonable performance just by matching the distributions of the source and target modality features. However, as previously discussed, having a small number of labeled samples improves performance significantly when compared to using unpaired samples in training.

\begin{table}[h]
  \caption{Comparison with different types of supervision, the baseline is a pre-trained \textit{Iresnet100}, rows with MMD corresponds to unpaired settings, and the last row is supervised with contrastive loss.}  \label{tab:training_arl_mmd}
  \centering
  \resizebox{\columnwidth}{!}{%

  \begin{tabular}{@{}lrrrrrr@{}}
  \toprule
  Architecture             & AUC   & EER   & Rank-1    & \begin{tabular}[c]{@{}c@{}} VR@\\FAR=0.1\% \end{tabular} \\ \midrule
  
  Baseline    & 94.55 & 12.73 & 31.33 & 14.00       \\ \hline
  \rowcolor{Gray}
  PDT + MMD (ip) & 94.02 & 13.33 & 52.67  & 40.00     \\
  \rowcolor{Gray}
  PDT + MMD (op) & 97.64 & 6.04  & 75.33  & 51.33  \\% just ZEI here
  \rowcolor{Gray}
  PDT + MMD (op + ip) & 99.49 & 3.33  & 90.00     & 78.67  \\ \hline
  
  %Iresnet100 (MMD, (Gen and ZEI pairs)) &98.91 &5.38 &86.00 &74.00 &84.67 &93.33 \\ 
  
  PDT + Contrastive         & 99.96 & 1.18  & 99.33 & 96.67    \\ \bottomrule
  \end{tabular}
  }
  \end{table}

\subsubsection{\textbf{Experiments with different Face Recognition models}}
%-----------------------------------------------------------------
% Redid
\begin{table}[h]
  \caption{Comparison with different architectures for the HFACE Task ( These are FaceX-Zoo models)}
  \label{tab:arch_arl}
  \centering
  \resizebox{\columnwidth}{!}{%
  \begin{tabular}{@{}lrrrrrr@{}}
  \toprule
  Architecture             & AUC   & EER   & Rank-1    & \begin{tabular}[c]{@{}c@{}} VR@\\FAR=0.1\% \end{tabular} \\ \midrule
  EfficientNet (baseline)  & 94.23 & 10.05 & 36.00    & 26.00        \\
  \rowcolor{Gray}
  EfficientNet + \sdt       & 99.79 & 2.73  & 94.67 & 84.00     \\
  MobileFaceNet (baseline) & 91.19 & 16.62 & 36.00    & 20.67       \\
  \rowcolor{Gray}
  MobileFaceNet + \sdt      & 99.76 & 2.69  & 93.33 & 79.33    \\
  ResNeSt (baseline)       & 97.41 & 10.00    & 62.67 & 36.00        \\
  \rowcolor{Gray}
  ResNeSt + \sdt            & 99.96 & 0.63  & 100.00   & 88.00         \\
  ResNet (baseline)        & 94.11 & 14.78 & 44.67 & 19.33       \\
  \rowcolor{Gray}
  ResNet + \sdt             & 99.95 & 0.74  & 96.67 & 86.67       \\
  TF-NAS (baseline)        & 93.93 & 13.33 & 38.67 & 26.67     \\
  \rowcolor{Gray}
  TF-NAS + \sdt             & 99.94 & 0.69  & 99.33 & 86.67       \\
  GhostNet (baseline)      & 90.67 & 18.67 & 33.33 & 20.00        \\
  \rowcolor{Gray}
  GhostNet + \sdt           & 99.96 & 1.22  & 98.67 & 94.00         \\
  HRNet (baseline)         & 90.89 & 16.67 & 36.00    & 22.00           \\
  \rowcolor{Gray}
  HRNet + \sdt              & 99.91 & 1.97  & 96.67 & 88.67       \\ \midrule
  Iresnet100 (baseline)    & 94.55 & 12.73 & 31.33 & 14.00        \\
  \rowcolor{Gray}
  Iresnet100 + \sdt         & 99.96 & 1.18  & 99.33 & 96.67       \\ \bottomrule
  \end{tabular}
  }
  \end{table}

  We have used \textit{Iresnet100} as the pre-trained FR model for all the experiments discussed in the previous section. In this section, we investigate whether the proposed approach generalizes to other FR architectures. We again perform these experiments in the ARL-VTF dataset using the $G\_VB0-P\_TB-$ protocol. For each experiment, we just switch the pre-trained FR model in the \sdt framework. In addition to \textit{Iresnet100}, we have used several other pre-trained FR models which were available publicly \footnote{\url{https://github.com/JDAI-CV/FaceX-Zoo}}.

  We first run these pre-trained models without the \sdt module to get a baseline performance. The results with the trained \sdt module are also shown in Table. \ref{tab:arch_arl}. From the table, it can be seen that the proposed approach works well with all the architectures used. The complexity and performance of each of these models are different. From these results, it is clear that the proposed approach can be used with any FR model architecture, given the \sdt module is trained together with it. Pre-trained FR models can thus be chosen for \textit{HFR} based on a tradeoff between accuracy and computational complexity.

\begin{table*}[t]
  \centering
  \caption{Rank-1 Accuracies for the cross-test between different architectures, rows are the architecture used for evaluation (A), and columns are the architecture 
  used for training the \sdt module (C), results for the same train-test architecture are highlighted in Gray.}
  \label{tab:mix match}
   \resizebox{0.99\textwidth}{!}{
\begin{tabular}{lccccccccc}
\toprule
{} &  AttentionNet(C) &  EfficientNet(C) &  GhostNet(C) &  HRNet(C) &  Iresnet100(C) &  MobileFaceNet(C) &  ResNeSt(C) &  ResNet(C) &  TF-NAS(C) \\
\midrule
AttentionNet(A)  &            \cellcolor{Gray} \AG{99.33} &            \AG{91.33} &        \AG{0.00} &     \AG{95.33} &          \AG{88.00} &             \AG{94.00} &      \AG{ 96.67} &      \AG{96.00} &     \AG{ 97.33} \\
EfficientNet(A)  &            \AG{70.67} &            \cellcolor{Gray} \AG{94.67} &        \AG{10.00} &     \AG{83.33} &          \AG{46.67} &             \AG{16.00} &      \AG{ 16.00} &      \AG{19.33} &     \AG{ 87.33} \\
GhostNet(A)      &            \AG{88.00} &            \AG{96.00} &        \cellcolor{Gray} \AG{98.67} &     \AG{92.67} &          \AG{81.33} &             \AG{92.67} &      \AG{ 89.33} &      \AG{92.00} &     \AG{ 94.67} \\
HRNet(A)         &            \AG{92.00} &            \AG{88.67} &        \AG{21.33} &     \cellcolor{Gray} \AG{96.67} &          \AG{82.00} &             \AG{94.00} &      \AG{ 93.33} &      \AG{90.00} &     \AG{ 97.33} \\
Iresnet100(A)    &            \AG{92.67} &            \AG{92.00} &        \AG{44.00} &     \AG{92.67} &          \cellcolor{Gray} \AG{99.33} &             \AG{60.00} &      \AG{ 80.67} &      \AG{84.67} &     \AG{ 84.67} \\
MobileFaceNet(A) &            \AG{87.33} &            \AG{81.33} &        \AG{49.33} &     \AG{86.00} &          \AG{80.67} &             \cellcolor{Gray}\AG{93.33} &      \AG{ 78.67} &      \AG{74.67} &     \AG{ 88.00} \\
ResNeSt(A)       &            \AG{95.33} &            \AG{93.33} &        \AG{1.33} &     \AG{96.67} &          \AG{82.00} &             \AG{92.67} &      \cellcolor{Gray} \AG{100.00} &      \AG{96.00} &     \AG{100.00} \\
ResNet(A)        &            \AG{99.33} &            \AG{96.67} &        \AG{20.00} &     \AG{97.33} &          \AG{94.00} &             \AG{95.33} &      \AG{ 98.67} &      \cellcolor{Gray} \AG{96.67} &     \AG{ 98.00} \\
TF-NAS(A)        &            \AG{92.00} &            \AG{92.00} &        \AG{42.67} &     \AG{92.67} &          \AG{90.00} &             \AG{85.33} &      \AG{ 88.00} &      \AG{87.33} &     \cellcolor{Gray} \AG{ 99.33} \\\bottomrule
\end{tabular}
   }

\end{table*}

\subsubsection{\textbf{Generalizability of the \sdt weights across architectures}}
%----------------------------------------------------------------------

In the previous sub-section, it has been shown that the proposed approach works with different FR models despite the differences in architecture and complexity. It is to be noted that, the architecture of the \sdt remains the same even though we used different FR architectures. We further investigated whether the \sdt module learned for one FR architecture would work for another. These experiments could provide insights into the learned transformations and could also indicate whether \sdt modules could be used with black-box models. For example, in Table. \ref{tab:mix match},  AttentionNet(C) (column)- EfficientNet(A) (row), means that the \sdt was added to AttentionNet model during training, the weights of the \sdt module trained with AttentionNet (checkpoint) was used in conjunction with EfficientNet architecture  for evaluation. From the results in Table. \ref{tab:mix match}, it can be seen that the proposed approach works well for most of the scenarios with the notable exception of GhostNet\cite{han2020ghostnet}, which could be due to GhostNet's unique architecture. GhostNet modules try to reduce the computational complexity of a network by reducing the filters needed to compute redundant feature maps. This is achieved by creating additional feature maps using cheap linear operations such as depth-wise separable convolutions on the output feature maps. The original feature maps and the newly created (ghost) features are concatenated in the ghost module to maintain the feature map sizes. We suspect that this operation makes the PDT learned together with GhostNet specifically optimized to the filters in the pre-trained backbone.  This results in the low performance of the \sdt module trained with the GhostNet module when it is used with other architectures. 
In summary, while the \sdt module works in most cases, the parameters of the \sdt module are ultimately learned in conjunction with a specific architecture, and it may not always generalize when used with black-box models. However, using multiple FR models to learn the \sdt weights in a distillation framework could improve the performance in the case of black-box FR models. This is left as future work and is beyond the scope of the current contribution.

\begin{figure*}[ht]
  \centering
  \includegraphics[width=0.99\linewidth]{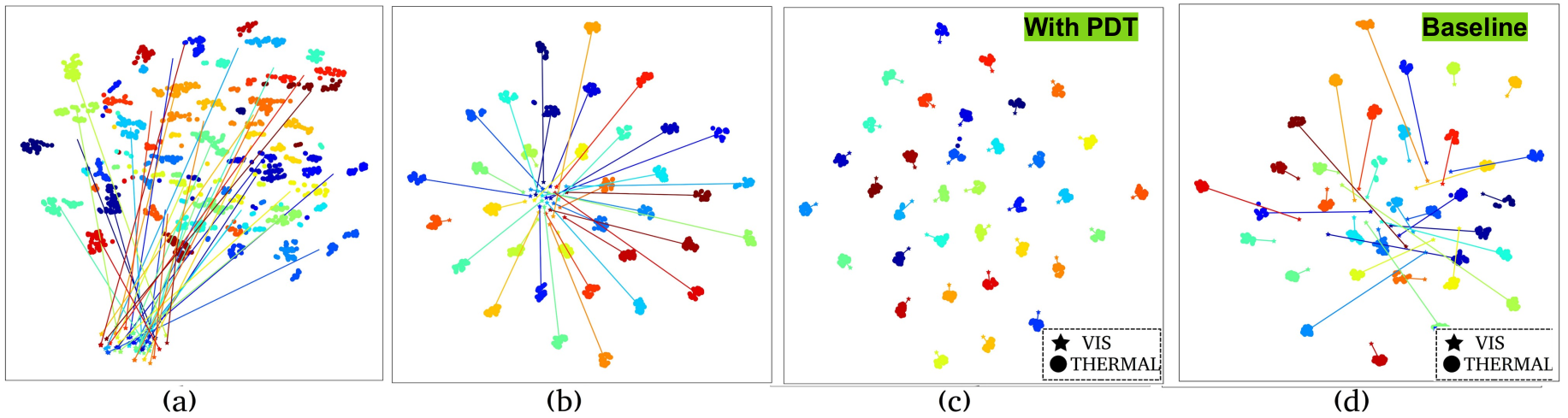}
  \caption{t-SNE plots for visible and thermal images from various stages, different colors indicate different identities. For the reference modality, only one cluster center is shown for clarity.} The lines connect the cluster center of visible and thermal images for each identity.  a) shows visible images and transformed thermal images in pixel space, b) from an intermediate feature map of the CNN, c) final embedding space with PDT, and d) final embedding space without PDT (baseline). It can be seen that the identities match in the final embedding space with the PDT block added.
  \label{fig:tsnec}
\end{figure*}
\subsubsection{\textbf{Visualization of the intermediate features}}
%----------------------------------------------------------------------

The t-SNE plots of thermal and visible images at various stages in the pipeline are shown in Fig. \ref{fig:tsnec}. The thermal and visual modalities are readily distinguished in the t-SNE plots at the input stage. The heterogeneous samples appear to form different distributions even after the \sdt stage. As we progress in the layers in the feature maps of the CNN, the distribution of thermal and visible features becomes more aligned. Towards the last layers and the final embedding output, the modalities align very well, i.e., the embeddings of the same identity in the thermal and visible spectrum grow closer in the embedding space, which is exactly what we need for the \textit{HFR} task. One important observation is that even after the \sdt block the distribution of features is disjoint.Despite the distribution disparity, the \textit{transformed} image following the \sdt block helps align the embeddings in the final layers.

\begin{figure}[h]
  \centering
  \includegraphics[width=0.99\linewidth]{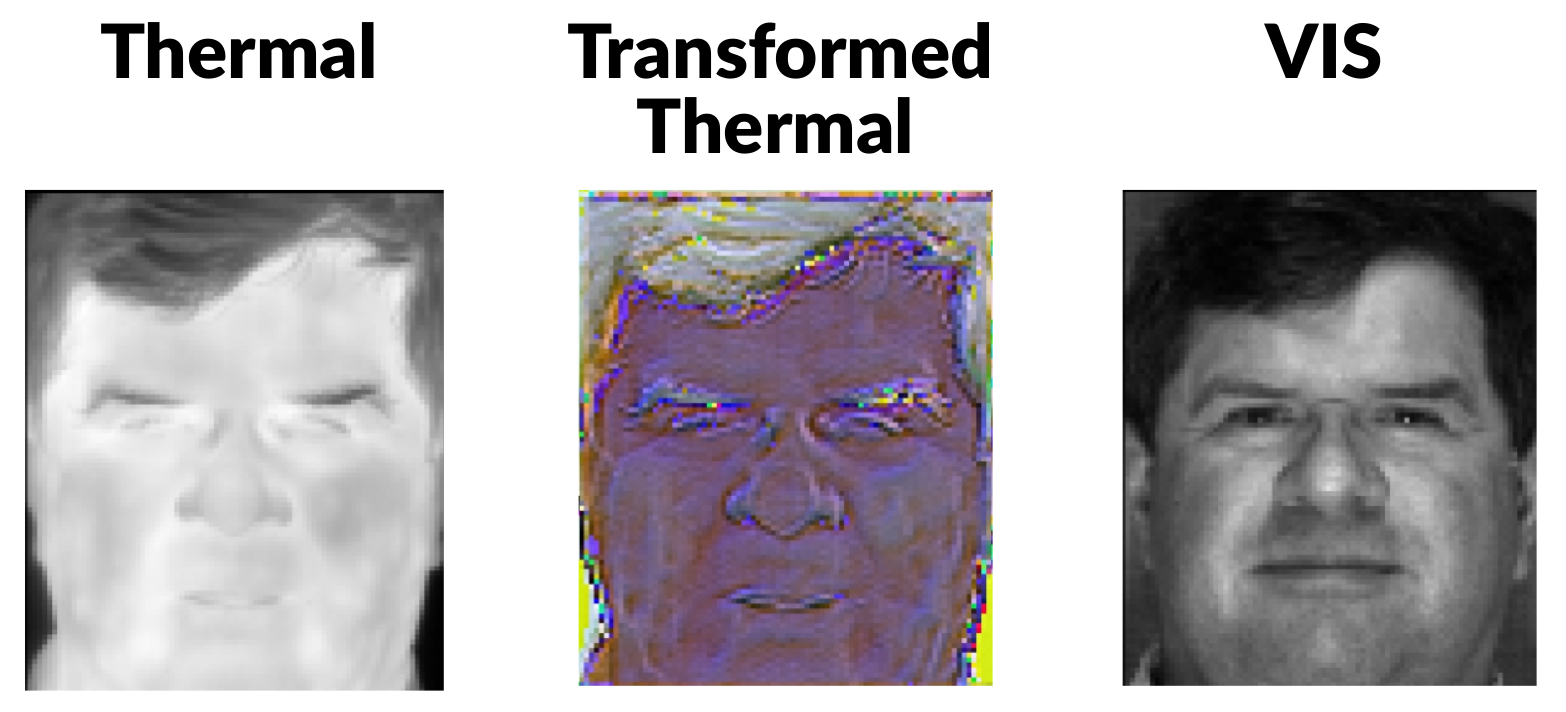}
  \caption{A visualization of thermal to vis \textit{HFR} scenario in Polathermal dataset, the \textit{Transformed-Thermal} is the intermediate output from the \sdt module, even though this image doesn't look visually similar to the VIS image, the embedding obtained from the \textit{transformed} image produces a high match score with the embedding extracted from the VIS image.}
  \label{fig:translated_image}
\end{figure}

Since the feature maps after the \sdt module are also three-channel images with the same resolution, they can be used to inspect the \sdt module's output. For example, Fig. \ref{fig:translated_image}, shows the thermal, visible, and \textit{transformed-thermal} image of the same person. In a broad sense, the face embeddings extracted from the \textit{transformed} images and visible spectrum images have a higher match score. Intuitively, one would expect the \textit{transformed} image to resemble the VIS image in order for this to occur; however, the results show that this is not the case; instead, the \textit{transformed} image only needs to preserve discriminative information for the \textit{HFR} task.

\section{Discussions}
\label{sec:discussions}
%============================================================================================================================================================

As opposed to the  computationally expensive synthesis based methods, the proposed approach introduces a simple, computationally simpler, plug and play module for heterogeneous face recognition that can be trained with a minimal number of samples. The proposed approach can be used even in scenarios where paired samples are not available. The \sdt approach achieves this by learning a parameter efficient transformation in the pixel domain that reduces the domain gap while retaining the discriminative information.

The proposed approach achieves state-of-the-art performance in many challenging benchmarks, as demonstrated by experiments in different thermal to VIS as well as other heterogeneous protocols. The new \sdt module added is parameter efficient and is generic enough for several heterogeneous scenarios.
As evidenced from the experiments in Table. \ref{tab:ablation_percentage}, the amount of training data required for our framework is very small, this makes it suitable for the real-world data starved heterogeneous settings. The approach was also found to perform well even in the absence of paired samples, this is of practical importance in heterogeneous scenarios where paired data is not readily available. The approach itself is modular and can be added to any pre-trained FR model. To summarize, we show that prepending a learnable neural network module to a pre-trained FR model yields state-of-the-art performance in a variety of challenging \textit{HFR} scenarios.

\subsection{Limitations and Future directions}

Currently, the architecture of the \sdt block is designed to be general so as to serve a wide range of heterogeneous scenarios. The current design of \sdt provides flexibility in terms of the receptive field due to the multi-branch architecture and CBAM module. This simple approach achieves comparable, and in many cases outperforms many computationally expensive state-of-the-art models. The receptive field of the branches in \sdt is a design parameter that can be optimized even further. It could be possible to optimize the architecture of \sdt for a specific heterogeneous scenario, which might boost the performance even further. Since the \proposed framework depends on the performance of the pre-trained network used in the framework, the overall performance would be impacted by the performance of the pre-trained network. For example, most of the standard pre-trained FR models struggle with extreme yaw angles and profile faces, meaning such a network would result in lower \textit{HFR} performance when there is extreme yaw angles present (as we see from the results in Tufts face dataset). Nevertheless, the proposed approach is generic enough to be applied to newer and robust face recognition models. Further, the \sdt approach is better suited for heterogeneous ``imaging'' modalities, as they share more structural similarities with visible face images. Heterogeneous modalities like drawn sketches do not satisfy this criterion and may be harder to adapt in the image domain. Methods like generative approaches might be better suited to these scenarios. Though the \sdt blocks trained with one architecture do not necessarily work well with other architectures, this could be enhanced to work with black-box models by employing multiple models in a teacher-student fashion. The proposed approach can also be combined with GAN-based generation methods, and can also be improved by using triplet or quadruplet loss functions. The proposed approach can be further extended with more tuning of the \sdt architecture with the likes of neural architecture search \cite{tan2019mnasnet}, training schedules, data augmentation, triplet training, and so on. However, as the title suggests, we hope that this approach will serve as a simple yet strong baseline motivating further research in heterogeneous face recognition.

\section{Conclusions}
\label{sec:conclusions}
%============================================================================================================================================================

In this work, we have proposed a simple yet effective framework for heterogeneous face recognition. Essentially, to convert a pre-trained FR model to an \textit{HFR} network we just prepend a novel neural network module for the target modality. The new module, called \proposed (PDT) is parameter efficient and does not require a lot of samples to train. Since the method performs well across a variety of FR architectures, it can be used to convert any face recognition model to a heterogeneous one. The design of the framework, training schedule, loss functions, and parameter selection are intentionally left simple to demonstrate the efficacy of the proposed approach. These can be further tuned to improve the results further. The proposed approach was found to outperform state-of-the-art methods in many challenging heterogeneous datasets. Further, we also introduce MCXFace heterogeneous face recognition dataset which contains multiple modalities which can be used for \textit{HFR} evaluations. Lastly, the source code, protocols, and datasets used are publicly available to make further extensions of the work possible.

\section*{Acknowledgment}

The authors would like to thank Innosuisse - Swiss Innovation Agency for supporting the research leading to results published in this paper.

\ifCLASSOPTIONcaptionsoff
  \newpage
\fi

% \clearpage

%\begin{spacing}{0.9}
\bibliographystyle{IEEEtran}
\bibliography{ref_translator}
% \end{spacing}

%

\begin{IEEEbiography}
  [{\includegraphics[width=1in,height=1.25in,clip,keepaspectratio]{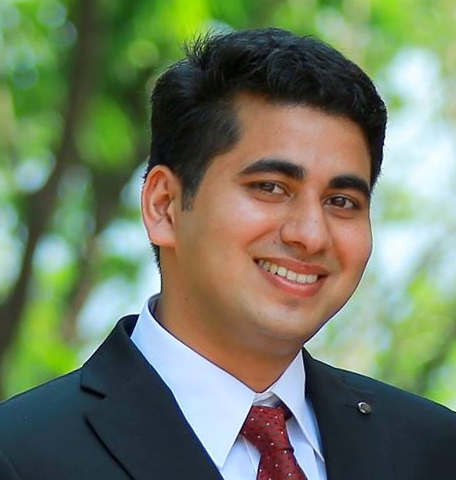}}]{Anjith George}
has received his Ph.D. and M-Tech degree from the Department of Electrical Engineering, Indian Institute of Technology (IIT) Kharagpur, India in 2012 and 2018 respectively. After Ph.D, he worked in Samsung Research Institute as a machine learning researcher. Currently, he is a research associate in the biometric security and privacy group at Idiap Research Institute, focusing on developing face recognition and presentation attack detection algorithms. His research interests are real-time signal and image processing, embedded systems, computer vision, machine learning with a special focus on Biometrics.
\end{IEEEbiography}

\begin{IEEEbiography}
  [{\includegraphics[width=1in,height=1.25in,clip,keepaspectratio]{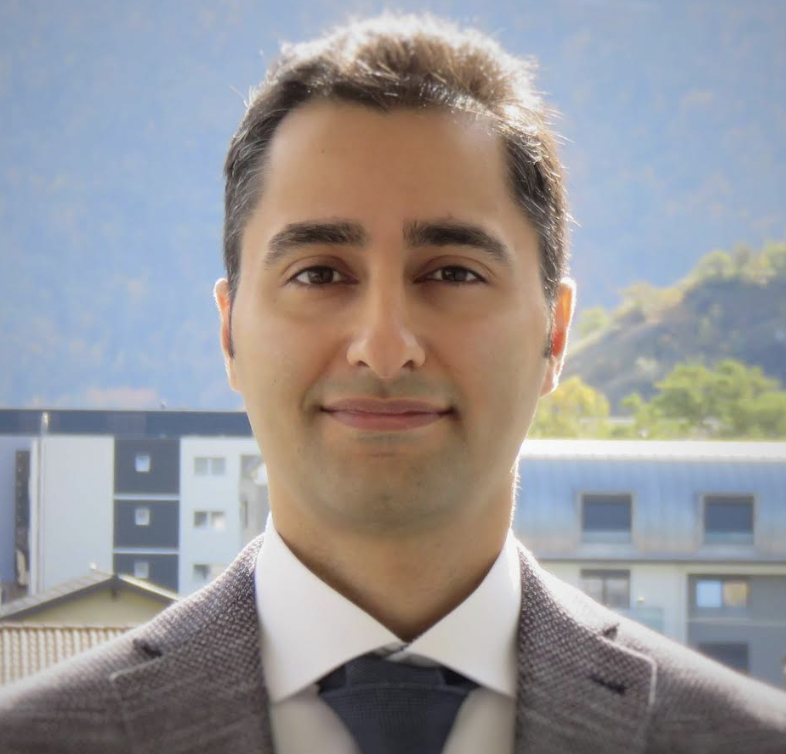}}]{Amir Mohammadi}
  obtained his PhD from EPFL, Switzerland in 2020 where he worked on face presentation attack detection and developed novel domain adaptation methods. He was a post-doctoral researcher at Idiap research institute where he worked on heterogeneous face recognition. Currently, he is working as senior data scientist at Eyeware working on head and gaze tracking.
\end{IEEEbiography}

\begin{IEEEbiography}
 [{\includegraphics[width=1in,height=1.25in,clip,keepaspectratio]{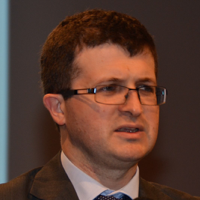}}]{S{\'e}bastien Marcel} heads the Biometrics Security and Privacy group at Idiap Research Institute (Switzerland) and conducts research on face recognition, speaker recognition, vein recognition, attack detection (presentation attacks, morphing attacks, deepfakes) and template protection. He received his Ph.D. degree in signal processing from Universit{\'e} de Rennes I in France (2000) at CNET, the research center of France Telecom (now Orange Labs). He is Professor at the University of Lausanne (School of Criminal Justice) and a lecturer at the  \'{E}cole Polytechnique F{\'e}d{\'e}rale de Lausanne. He is also the Director of the Swiss Center for Biometrics Research and Testing, which conducts certifications of biometric products..
\end{IEEEbiography}
\vfill

\end{document}